\documentclass{article}

 \usepackage[main, final]{neurips_2026}

\usepackage[utf8]{inputenc} 
\usepackage[T1]{fontenc}    
\usepackage{hyperref}       
\usepackage{url}            
\usepackage{booktabs}       
\usepackage{amsfonts}       
\usepackage{nicefrac}       
\usepackage{microtype}      
\usepackage{xcolor}         
\usepackage{graphicx}
\usepackage{subcaption}
\usepackage{amsmath}
\usepackage{bbm}
\usepackage{hyperref}
\usepackage{xurl}
\usepackage{wrapfig}
\usepackage{natbib}
\usepackage{booktabs}
\usepackage{hyperref}   
\usepackage{placeins}

\usepackage{amsmath}
\usepackage{amssymb}
\usepackage{mathtools}
\usepackage{amsthm}
\usepackage{float} 
\usepackage{todonotes}
\usepackage{multirow}

\theoremstyle{plain}

\theoremstyle{definition}

\theoremstyle{remark}

\newcommand{\model}{f_\theta}

\newcommand{\train}{D_{\text{train}}}
\newcommand{\test}{D_{\text{test}}}

\title{Estimating Model-Level Membership Inference Vulnerability Without Reference Models}

\author{%
  Euodia Dodd \\
  Imperial College London\\
  \texttt{e.dodd24@imperial.ac.uk} \\
  \And
  Nata\v{s}a Kr\v{c}o \\
  Imperial College London\\
  \texttt{n.krco23@imperial.ac.uk} \\
  \And
  Igor Shilov \\
  Imperial College London\\
  \texttt{i.shilov23@imperial.ac.uk} \\
  \And
  Matthew Wicker \\
  Imperial College London\\
  \texttt{m.wicker@imperial.ac.uk} \\
  \And
  Yves-Alexandre de Montjoye \\
  Imperial College London\\
  \texttt{demontjoye@imperial.ac.uk} \\
}

\begin{document}

\maketitle

\begin{abstract}
  Membership inference attacks (MIAs) have emerged as the standard tool for evaluating the privacy risks of AI models. However, state-of-the-art attacks require training numerous, often computationally expensive, reference models, limiting their practicality. We present a novel approach for estimating model-level vulnerability to the Likelihood Ratio Attack (LiRA), the strongest available attack, directly from the train and test loss distributions of the target model and without training any reference models. We show that LiRA's per-sample signal decomposes into a variance-ratio term and a residual mean-shift term, with the relative contribution of each determined by how much training collapses model uncertainty at the trained sample. This places models on a continuum, with different regimes calling for different reference-free loss-based statistics as proxies for LiRA TPR. The shapes of the loss distributions themselves indicate which proxy applies. We instantiate the framework with two natural proxies. At the heavy-tailed end, the LOSS attack TNR predicts LiRA TPR@FPR=$10^{-3}$ with RMSE 0.036 across 10 image classification architectures and 4 datasets, outperforming low-cost reference-model attacks such as RMIA. At the symmetric end, the LOSS attack AUC predicts LiRA TPR with RMSE 0.018 across five GPT-2 sizes from 10M to 1B parameters. 
\end{abstract}

\section{Introduction}
\label{sec:introduction}
\looseness=-1
Large-scale machine learning systems have seen rapid adoption across healthcare, finance, legal, and other domains, 
with many models now being fine-tuned on sensitive and confidential data. A large body of research has 
shown that these models may inadvertently memorize and leak 
samples within their training data (often outliers)~\citep{carlini2022onion, carlini2019secret}, raising strong privacy concerns.

Membership inference attacks (MIAs) have become the primary tool to measure the empirical privacy leakage of a model, providing both a measure of practical risk and establishing a bound on the success of other types of attacks~\citep{ponomareva2023dpfy}. The privacy risk of a model is quantified by the attack's True Positive Rate (TPR) at a low False Positive Rate (FPR)~\citep{carlini2022membership, ye2022enhanced, zarifzadeh2023low, watson2021importance}, focusing on the members an attack can confidently infer. This is also aligned with legal requirements, such as the EU GDPR ``reasonably likely'' standard for singling out and the UK Information Commissioner's Office's recent guidance~\citep{edpb2024opinion28, ICO2025SecurityDataMinimisation}.

\subsection{Problem Statement}
\looseness=-1
The most effective MIAs, such as the Likelihood Ratio Attack (LiRA)~\citep{carlini2022membership}, provide an estimate of the upper bound of the privacy risk of a model. However, they require training tens to hundreds of reference models, each requiring computational resources equivalent to the original model. 
It has also been shown that the success of an attack increases with the number of reference models, with recent work showing continuous improvement with 256 reference GPT-2 models~\citep{hayes2025strongmembershipinferenceattacks}, and that reference models need to be trained with the same architecture and setup as the model under attack to be optimal~\citep{carlini2022membership, zarifzadeh2023low}. The high computational cost of SOTA attacks limits their practicality as privacy risk evaluation tools, especially for complex models and LLMs. Importantly, this cost becomes even more prohibitive in realistic workflows such as hyperparameter and architecture selection, where practitioners must compare privacy-utility tradeoffs across many training configurations~\citep{ponomareva2023dpfy}, as well as continuous monitoring of routinely retrained production models and rapid pre-deployment risk assessments. 

A recent line of research has aimed to decrease the number of reference models needed to perform attacks. The state-of-the-art for this, RMIA, however still requires the training of at least one reference model~\citep{zarifzadeh2023low}. Instead, another line of research focused on methods to score the vulnerability of individual training samples to MIAs. These however do not provide an overall model-level risk metric~\citep{pollock2025free, leemann2024my}.

\textbf{We here propose a new framework to estimate the vulnerability of a model to state-of-the-art membership inference attacks directly from the shape of its train and test loss distributions, without training any reference models.}

\subsection{Our Approach}

Membership inference attacks take advantage of the fact that models tend to be more confident and exhibit lower loss in their predictions for data samples they have seen during training. This differential behavior is a direct consequence of the optimization process that aims to minimize the loss. Some samples are more difficult to learn than others due to their difficulty, rarity or position as outliers in the dataset distribution, and are prone to being memorized. 

We focus on LiRA as it represents, to our knowledge, the strongest membership inference attack currently available for both classical deep learning models and language models~\citep{hayes2025strongmembershipinferenceattacks}. Estimating LiRA's TPR@low FPR therefore provides a tight estimate of the practical privacy risk from the strongest known attack. While we acknowledge that the attack landscape continues to evolve, focusing on the strongest attack allows for reproducible and interpretable auditing, and for tracking progress in the future as stronger attacks are developed.

We argue that LiRA's per-sample signal decomposes into two components (Section \ref{sec:intuition}): one that captures training-induced collapse of the model's predictive uncertainty, and one that captures a population-level mean shift. The relative importance of these two components varies continuously across models and datasets, placing models on a continuum of uncertainty collapse. At one end, training strongly reduces uncertainty for training (member) samples, producing a heavy-tailed loss distribution for non-members; at the other, uncertainty remains largely unchanged, and only a residual mean-shift signal remains. 

\looseness=-1
Importantly, this continuum is directly observable from the shape of the loss distributions. In the strong-collapse regime, the absence of a heavy tail in the training distribution reflects a population of samples that have migrated from high loss (had they not been seen) to low loss through training. In the weak-collapse regime, where distributions remain symmetric, the signal reduces to a population-level mean separation.


We also empirically confirm both ends across a wide variety of setups: Across 9 image classification architectures on 4 datasets, image classifiers sit near the heavy-tailed end. Across five GPT-2 models from 10M to 1B parameters, LLMs sit near the symmetric end.

Leveraging these insights, we propose a family of loss-based proxies for model-level MIA risk, computed from readily available train and test set loss distributions. In the strong-collapse regime, tail-based statistics capture the fraction of samples that have undergone tail-to-head migration, while in the weak-collapse regime, mean-separation statistics capture the remaining signal. The shape of the loss distribution itself serves as a diagnostic for selecting the appropriate estimator.

We instantiate the framework with two natural proxies, the LOSS attack TNR and the LOSS attack AUC, each appropriate at a different end of the loss-distribution-shape spectrum. We show that, in the strong-collapse regime, LOSS TNR (a tail statistic) provides an accurate estimate of LiRA TPR, while in the weak-collapse regime, the LOSS AUC (a mean-separation statistic) is more appropriate.

This work targets developer-side privacy auditing. The practitioner has access to the target model and to both its training and held-out test sets, and wants a low-cost estimate of how vulnerable the model would be to a state-of-the-art MIA mounted by an external adversary. 

\section{Preliminaries}
\label{sec:preliminaries}
\subsection{Notation}

We denote with $\model:\mathcal{X}\rightarrow\mathcal{Y}$ a machine learning classification model parameterized by $\theta$, with input space $\mathcal{X}$ and output space $\mathcal{Y} = [0,1]^n$. Model $\model\leftarrow \mathcal{T}(\train)$ is trained on dataset $\train\sim \mathcal{X}\times\mathcal{Y}$ 
using algorithm $\mathcal{T}$ and loss function $\ell$. A separate test dataset $\test\sim \mathcal{X}\times\mathcal{Y}$ is reserved for evaluation and is never used during training. For sample $(x, y)$, we write $\model(x)_{y}$ for the predicted probability for class $y$ and $\ell(x,y) = \ell(\model(x)_{y}, y)$ for the corresponding loss. 

Membership inference attacks (MIAs) aim to determine whether a given record $(x,y)$ was used to train target model $\model$. Specifically, given target model $\model$ and target record $(x,y)\in\mathcal{X}\times\mathcal{Y}$, an MIA $A(x,y, \model)$ aims to infer whether $(x,y)\in\train$ ($(x,y)$ is a \textit{member}) or not ($(x,y)$ is a \textit{non-member}). We denote the target model's loss on $(x,y)$ with $\ell_{\text{target}}(x,y)$. The attack outputs a membership score that quantifies the likelihood of the target sample $(x,y)$ being part of the training data of the target model. The attacker then makes a binary membership prediction according to threshold $\tau$:

\begin{equation}
A_{\tau}(x, y ,\model) = \begin{cases}
1 & \text{if } A(x, y ,\model) \geq \tau \\
0 & \text{otherwise}
\end{cases}
\end{equation}

A simple and widely used example is the LOSS attack $A^{\text{LOSS}}$~\citep{yeom2018privacy}, where a sample's membership score is the loss of the target model on it and $\mathbbm{1}$ is the indicator function:

\begin{equation}
    A^{\text{LOSS}}_{\tau}(x,y) = \mathbbm{1}[-\ell_{\text{target}}(x,y) > \tau]
\end{equation}

We report the model-level risk as the True Positive Rate (TPR) at a fixed False Positive Rate (FPR)=$\alpha$~\citep{carlini2022membership, zarifzadeh2023low, ye2022enhanced} of an attack.

SOTA attacks rely on \emph{reference models}, models trained on auxiliary datasets $D^i_{\text{reference}}\sim \mathcal{X}\times\mathcal{Y}$ using the same architecture and training process as the target model. For a sample $(x,y)$, an ``OUT'' reference model $f^i_{\theta_{OUT}}$ is trained on data excluding $(x,y)$, while an ``IN'' reference model $f^i_{\theta_{IN}}$ is trained with it. This distinction enables the attacker to compare model behavior in the presence and absence of the target sample, improving its power. For a sample $(x,y)$ and a dataset $D^i_{\text{reference}}$, we denote with $\ell^i_{\text{OUT}}(x,y)$ the loss of an OUT model on $(x,y)$, and analogously for an IN model. For K IN reference models, we define the mean loss:

\begin{equation}
    \Bar\ell_{in}(x,y) = \frac{1}{K} \sum_{k=1}^K \ell(f_{\text{IN}}^k(x, y))
\end{equation}
and analogously for $\Bar\ell_{out}(x,y)$.

\subsection{SOTA Reference Model-Based Membership Inference Attacks}

The current state-of-the-art reference-model based attack is the Likelihood Ratio Attack (LiRA)~\citep{carlini2022membership}. It estimates the loss distributions of models trained with and without a target sample $(x,y)$, and computes their likelihood ratio as a membership score. In the \textit{online} attack, the attacker trains both ``IN'' and ``OUT'' reference models, and in the more efficient \textit{offline} version, only trains ``OUT'' models. We adopt the strongest online attack in our experiments.

\looseness-1
Recent work has explored methods to reduce the number of reference models needed for an effective attack. The state-of-the-art approach in this setting is RMIA~\citep{zarifzadeh2023low}. It computes membership scores by comparing the target sample to samples drawn from the population using simple pairwise tests, and has been shown to outperform LiRA when few reference models are available. We compare our method to it.

\section{Intuition}
\label{sec:intuition}
\paragraph{Why the tail separates members from non-members.}

\begin{figure*}[h]
    \centering
    \includegraphics[width=0.9\linewidth]{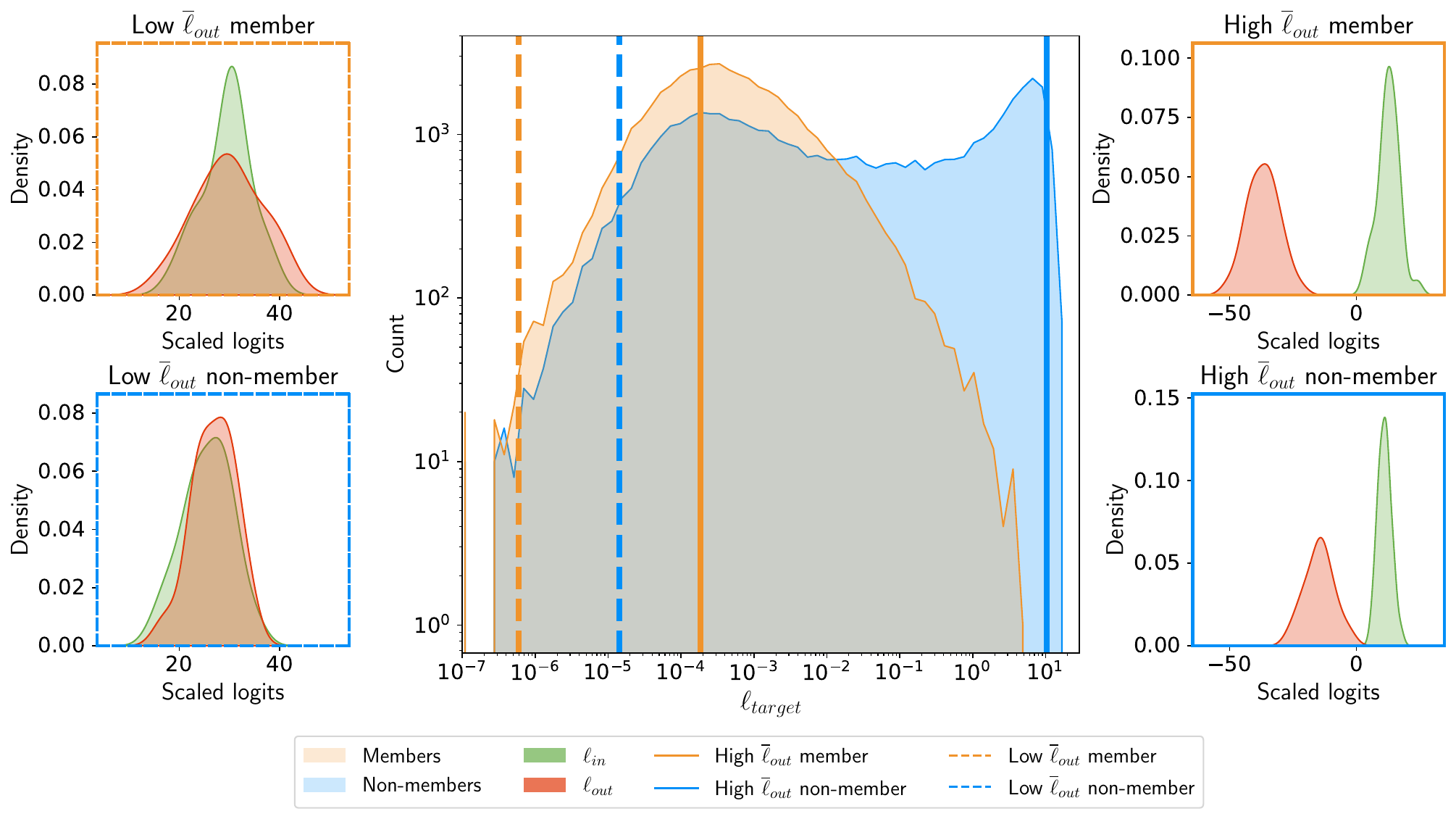}
    \caption{
    \textbf{Vulnerable members are mostly those that moved from tail to head.} The central panel shows member (orange) and non-member (blue) loss distributions on the target model. \textbf{Non-members dominate the high-loss tail} (right side), while members concentrate at low loss. Vertical lines mark $\ell_{target}$ values for representative sample groups. Side panels display the IN and OUT scaled logit distributions that LiRA uses to compute membership scores for these samples: left panels show samples with low $\ell_{out}$, right panels show samples with high $\ell_{out}$. In particular, note how the \textbf{LiRA-vulnerable members} (orange solid line, top-right panel) have low $\ell_{target}$ but high $\ell_{out}$ indicating these samples would naturally fall in the high-loss tail but moved to the low-loss $\ell_{target}$ head through training. Results shown are for ResNet-18 trained on CINIC-10.}
    \label{fig:intuition}
\end{figure*}

Figure~\ref{fig:intuition} shows the member and non-member $\ell_{\text{target}}$ distributions on a semilog scale for ResNet-18 on CINIC-10; complete results are shown in Appendix \ref{app:dists}. In both cases, most probability mass lies at low loss (the ``head''), with a smaller fraction at high loss (the ``tail''). For members, the head corresponds to learned or memorized examples, while the tail comprises harder-to-learn samples~\citep{pollock2025free}. For non-members, the head reflects samples the model generalizes to, whereas the tail contains difficult or outlier cases that the model fails to fit. The two distributions overlap heavily at the head, and what separates them is the tail, which is dominated by non-members. Training has shrunk the member tail by pushing high-loss examples to low loss.






\paragraph{Vulnerable members are those that are missing from the tail.}


Assuming i.i.d. sampling, the absence of a heavy tail for members suggests samples that would have been high-loss (if unseen) have shifted to the low-loss region after training. These tail-to-head samples are now difficult to distinguish from easy non-members, but are exactly those that state-of-the-art membership inference attacks such as LiRA target. Figure~\ref{fig:intuition} supports this hypothesis: samples with the highest LiRA scores exhibit low ${\ell}_{target}$, yet high $\overline{\ell}_{out}$, aligning with the heavy tail of the non-member distribution (see Appendix~\ref{app:dists} for results for all setups). This pattern indicates that training shifted these examples from the tail to the head. This observation is not specific to a particular attack, but reflects a general property of how training reshapes the loss distribution.


\paragraph{When tail-to-head migration occurs.}

Tail-to-head migration requires that training pin down the model's behavior at the trained sample: a sample's loss becomes reliably low and distinguishable from a non-member at the same $x$ only when including $(x, y)$ in training meaningfully changes how the model behaves at $x$. When the training set carries heavy near-duplication, the model's behavior at $x$ is already determined by similar examples elsewhere, and including a specific $(x, y)$ does not change it. Section~\ref{sec:gp-analysis} formalizes this and shows that real models lie on a continuum determined by how much training pins down behavior on individual samples. The signal that an MIA can extract differs at the two ends of this continuum, and so does the appropriate reference-free proxy.

    


\subsection{Theoretical Analysis via Gaussian Process Regression}
\label{sec:gp-analysis}

We study membership inference attacks against Gaussian Process (GP) regression models, which, while simpler than deep networks, provide an analytically tractable setting to examine the relationship between LiRA and LOSS attacks. This setting is particularly instructive as it provides a closed-form solution for the LiRA attack—specifically, the performance of LiRA as the number of reference models tends towards the number of data points. Let $\mu_{\mathcal{D}}(x)$ and $\sigma_{\mathcal{D}}^2(x)$ denote the GP’s predictive mean and variance, respectively. We denote the corresponding properties for members as $\mu_{\text{mem}}$ and $\sigma_{\text{mem}}$, and for non-members as $\mu_{\text{non}}$ and $\sigma_{\text{non}}$. Under the GP, the likelihood ratio of LiRA, $\Lambda(s) = \ln \left( p(y | x, H_{mem})/p(y | x, H_{non})\right)$, is computed by substituting the Gaussian log-density:

\begin{equation}
\label{eq:lira-full}
\Lambda
= \underbrace{\tfrac{1}{2}\ln\!\left(
      \frac{\sigma^2_{\text{non}}}{\sigma^2_{\text{mem}}}
  \right)}_{\text{variance-ratio term}}
+ \underbrace{\frac{(y-\mu_{\text{non}})^2}{2\sigma^2_{\text{non}}}
  - \frac{(y-\mu_{\text{mem}})^2}{2\sigma^2_{\text{mem}}}
  }_{\text{residual term}}.
\end{equation}

LiRA's advantage over LOSS, which thresholds the squared error
$\ell(x,y) = (y - \mu_{\mathcal{D}}(x))^2$, lives in the variance-ratio term i.e the per-sample uncertainty signal. Its size is dependent on how much $\sigma^2$ collapses on the training sample, giving two limits. 

\textbf{Strong-collapse limit $\sigma^2_{\text{mem}}(x) \ll \sigma^2_{\text{non}}(x)$.} In the
interpolating limit ($\mu_{\text{mem}} \to y$,
$\sigma^2_{\text{mem}} \to \sigma^2_\epsilon$),
Eq.~\eqref{eq:lira-full} reduces to an uncertainty-calibrated
residual:
\begin{equation}
\label{eq:lira-regime1}
A_{\text{LiRA}}
= \mathbbm{1}\!\left[\,
    \frac{(y-\mu_{\text{non}})^2}{2\sigma^2_{\text{non}}}
  + \frac{1}{2}\ln\!\left(
      \frac{\sigma^2_{\text{non}}}{\sigma^2_\epsilon}
    \right)
  > \tau_{\text{lira}}
  \,\right].
\end{equation}

Samples with high  $\sigma^2_{\text{non}}$ receive a large LiRA bonus. These samples are exactly the tail-to-head population. At the population level this produces heavy-tailed non-member loss distributions, and LOSS TNR at low FNR (the proportion of non-members correctly classified as such by the LOSS attack), which counts exactly this migrated population, is the natural summary. We provide more details on the strong collapse regime in Appendix~\ref{app:strong-collapse}.

\textbf{Weak-collapse limit $\sigma^2_{\text{mem}}(x) \approx \sigma^2_{\text{non}}(x) \equiv
\sigma^2(x)$.}  The variance-ratio term vanishes and Eq.~\eqref{eq:lira-full} reduces to a mean-shift:

\begin{equation}
\label{eq:lira-regime2}
\Lambda \approx
\frac{(\mu_{\text{mem}} - \mu_{\text{non}})
      \bigl(2y - \mu_{\text{mem}} - \mu_{\text{non}}\bigr)}
     {2\sigma^2(x)}.
\end{equation}

LiRA can now only detect shifts in the model's mean prediction under training. LOSS uses the same signal at the population level, but LiRA retains an advantage as it measures the per-sample mean shift directly via $\mu_{\text{non}}$ from reference models, while LOSS can only threshold the target model's raw loss. This limit produces near-symmetric loss distributions, and LOSS AUC (a measure of population-level mean separation) is the natural summary.

\textbf{Real models live between these limits.} Eqs. \ref{eq:lira-regime1} and \ref{eq:lira-regime2} are the endpoints of a continuum, not a classification of models. For any real target, $\sigma^2_\text{mem}/\sigma^2_\text{non}$ varies across samples and sits somewhere between 0 and 1, with the population-level distribution of this ratio determining where the model falls along the axis. As the ratio shrinks on average, the variance-ratio term contributes more to $\Lambda$, the non-member loss distribution develops a heavier tail, and TNR becomes a stronger proxy for LiRA TPR than AUC. As the ratio grows toward 1, the residual term carries the signal, the loss distribution becomes more symmetric, and AUC overtakes TNR. Both proxies are reference-free, and the loss-distribution shape, itself reference-free, tells the practitioner which one is more appropriate. We provide a more detailed analysis of the efficacy of both summary statistics as proxies under different regimes in Appendix \ref{app:proxy-comparison}. 

\begin{figure}[h!]
    \centering
    \includegraphics[width=0.8\linewidth]{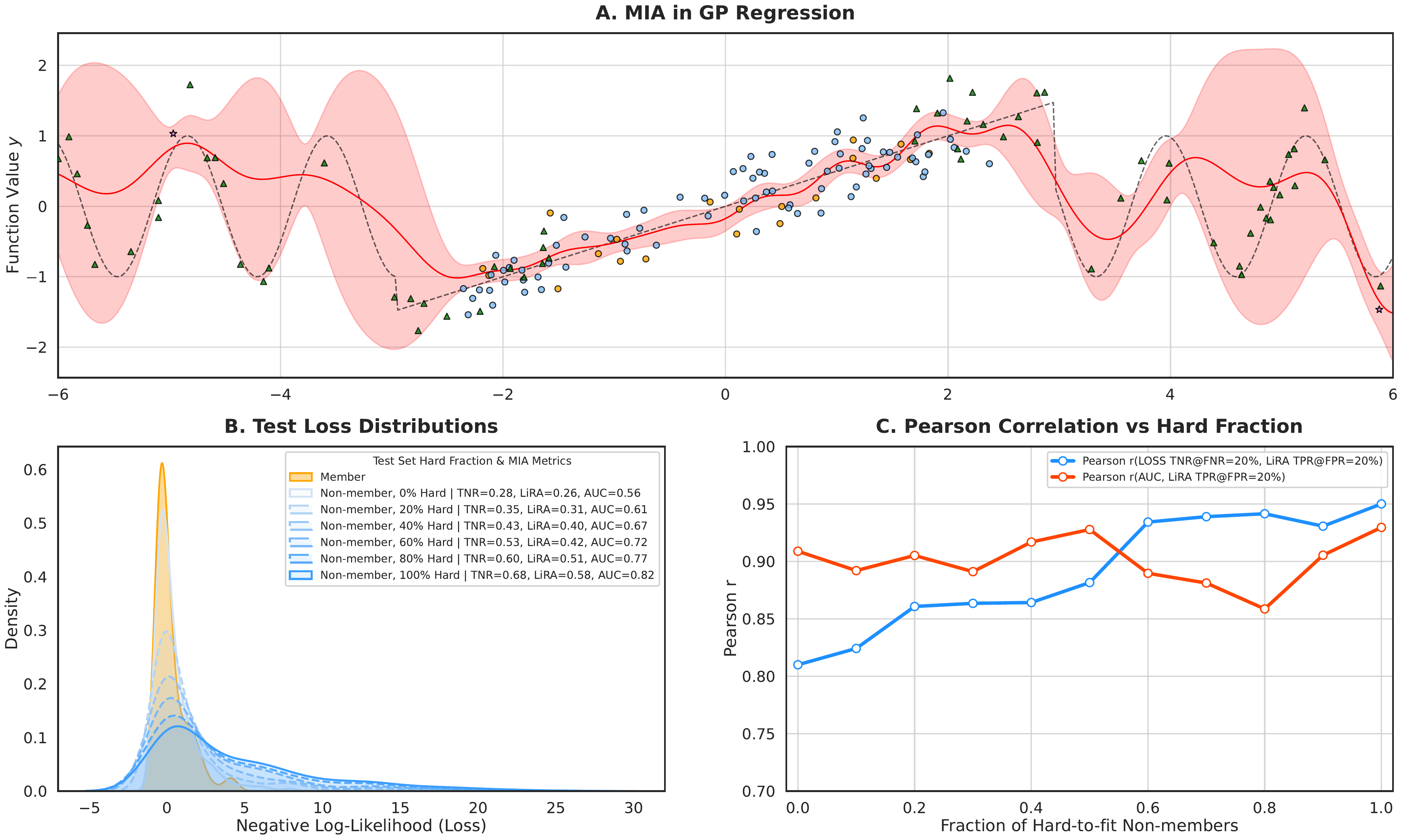}
    \caption{\textbf{Relationship between loss-distribution shape
    and the best loss-based proxy for model-level MIA risk.}
    \textbf{(A)} Toy GP regression setting with easy-to-fit and
    hard-to-fit members (orange) and non-members (blue), showing how
    sparse, high-uncertainty regions induce large losses for hard
    non-members. \textbf{(B)} As the fraction of
    hard-to-fit non-members increases, the non-member loss
    distribution becomes increasingly heavy-tailed, increasing both
    LiRA TPR and the predictive power of tail-sensitive metrics.
    \textbf{(C)} Pearson correlation with LiRA TPR shows
    that LOSS AUC is the stronger predictor when the loss
    distributions are relatively symmetric, whereas LOSS TNR
    overtakes it once the non-member distribution is sufficiently
    heavy-tailed.}
    \label{fig:gp-toy}
\end{figure}

\textbf{The continuum is visible in a controlled toy.} We illustrate this directly using a piece-wise regression problem where ``easy'' examples (linear) occupy the center and ``hard'' examples (non-linear) form the tails. As shown in Figure~\ref{fig:gp-toy}A, we purposefully sample fewer training points from the tails to induce high epistemic uncertainty, reflecting the conditions for tail-to-head migration.

By varying the fraction of hard-to-fit non-members continuously from $0\%$ to $100\%$, we slide the model along the continuum from the no-collapse end toward the full-collapse end. As this fraction
increases, the non-member loss distribution shifts from symmetric to
increasingly heavy-tailed (Figure~\ref{fig:gp-toy}B),
tracing the path from the weak-collapse limit to the strong-collapse limit. The bottom-right panel
shows the consequence: Pearson correlation between LOSS AUC and
LiRA TPR is highest when the distribution is symmetric, while LOSS
TNR overtakes it once the tail develops. We empirically show this to also be the case for image models and report results in Appendix~\ref{app:relaxloss}.

\section{Experimental Setup}
\label{sec:setup}
\paragraph{Datasets.} We select 4 image classification datasets widely used in the literature: MNIST, CIFAR-10, CINIC-10~\citep{darlow2018cinic} and CIFAR-100~\citep{krizhevsky2009learning}. MNIST contains 70,000 grayscale 28×28 images of handwritten digits (0-9). CIFAR-10 has 60,000 32×32 color images across 10 natural object classes, and CINIC-10 combines CIFAR-10 with downsampled ImageNet data~\citep{5206848}, creating 270,000 32×32 images across the same 10 classes as CIFAR-10 but with greater data volume and diversity. Finally, CIFAR-100 contains the same images as CIFAR10, but is divided into 100 fine-grained classes. 

\paragraph{Models.} We train 10 neural network architectures commonly used for image classification, ranging between 270K and 86M parameters on all 4 datasets. We train ResNet-20~\citep{he2016deep}, WRN28-2~\citep{zagoruyko2016wide}, MobileNetV2~\citep{sandler2018mobilenetv2}, DenseNet121~\citep{huang2017densely}, WRN40-4, ResNet-18, WRN28-10, VGG11 and VGG16~\citep{simonyan2015very}. Additionally, we finetune ViT-B-16~\citep{dosovitskiy2020image} pretrained on ImageNet on CIFAR-10. 


\paragraph{Training.} 
We follow the setup from~\citet{carlini2022membership} where each model is trained on a randomly selected subset consisting of 50\% of the samples from the training split as members, with the other 50\% reserved as non-members. We use SGD, momentum 0.9, with a cosine annealing learning rate schedule and a batch size of 256. Models are trained for 50 to 200 epochs using weight decay and data augmentation to achieve a high training accuracy of over 90\%.

\paragraph{Attacks.} We instantiate LiRA and RMIA in the online setting, and follow the ``mirror'' setup from~\citet{carlini2022membership} with 2 augmentations. For the image models, we use $K=64$ reference models (32 IN and 32 OUT). For the GPT-2 models, we follow the implementation from~\citet{hayes2025strongmembershipinferenceattacks}, using $K=127$ reference models. RMIA is instantiated with $K=2$ reference models. For the LOSS attack, we collect per-sample loss values at the final epoch of training for the target model. Appendix~\ref{app:cost} reports the computational cost of each attack.

\section{Estimating Vulnerability to SOTA MIAs}
\label{sec:method}
The GP analysis in Section \ref{sec:intuition} predicts that two reference-free metrics track LiRA TPR depending on where the target model sits on the variance-collapse continuum: TNR at the heavy-tailed end, where the variance-ratio term dominates and tail-to-head migration is the operative mechanism, and AUC at the symmetric end, where only the residual mean-shift signal remains. We now confirm both empirically. 


\subsection{Image Recognition Models}

Image classifiers are expected to lie in the strong-collapse regime, where membership inference signal is concentrated in the tail. Consistent with the theory, we observe a heavy-tailed non-member loss distribution, suggesting that tail statistics should govern attack performance.

The TNR measures how good the LOSS attack is at confidently identifying non-members, providing a measure of how ``separable'' the tail is from the member distribution. For a set of non-members $D_{\text{test}}$, model $f_\theta$ and threshold $\tau$, we compute the LOSS TNR as the fraction of correctly identified non-members:


\begin{equation}
    \text{TNR}_{\text{LOSS}}(D_{\text{test}}, f_{\theta}, \tau)  = \frac{  \sum_{(x,y)\in D_{\text{test}}} A^{\text{LOSS}}_\tau(x, y, f_\theta)  }{|D_{\text{test}}|}
\end{equation}

We select $\tau$ to achieve a False Negative Rate equal to the FPR of the LiRA attack for which we are estimating the TPR. In line with the literature, we focus on FPR=0.001, but present results for different FPR values (see Appendix \ref{app:metrics_fprs}). 

We select FNR=FPR as the most natural choice, but show our method is robust to the choice of FNR threshold: as shown in Appendix \ref{app:sensitivity}, LOSS TNR maintains strong predictive performance across a wide range of FNR values, with RMSE remaining relatively stable.

\begin{wrapfigure}{r}{0.5\textwidth}
  \begin{center}
    \vspace{-20pt}
    \includegraphics[width=0.45\textwidth]{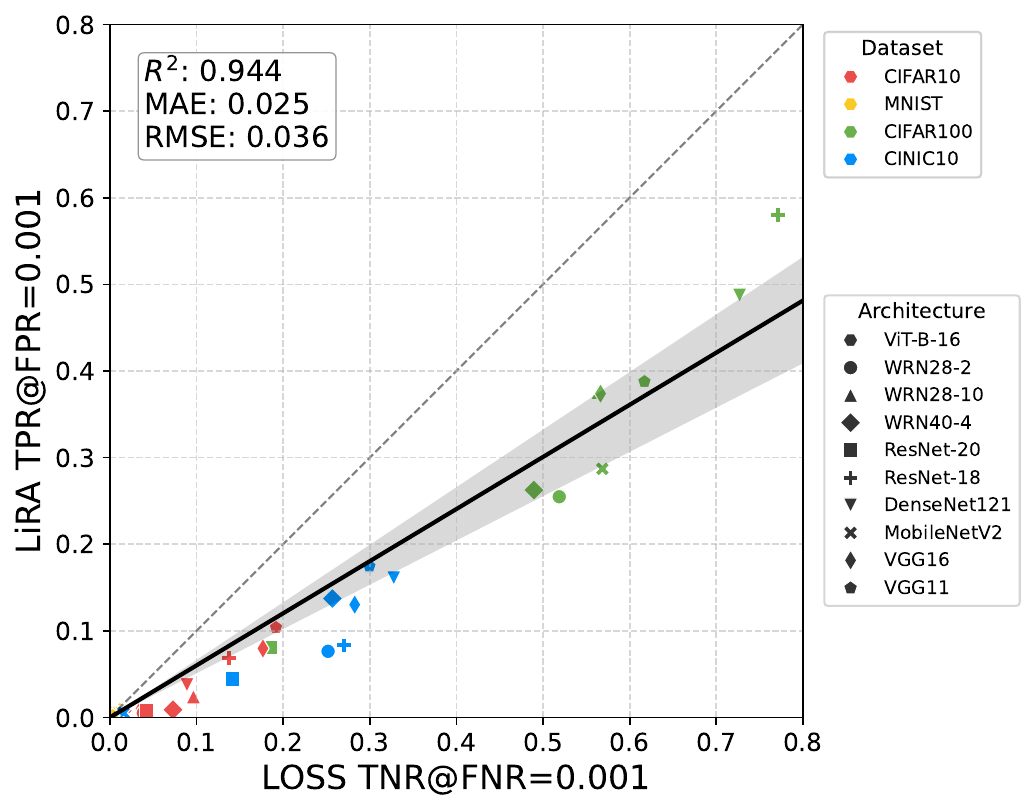}
  \end{center}
  \caption{LOSS TNR reliably predicts LiRA TPR@FPR=0.001 across varied model architectures and datasets. The linear relationship (solid black line) exhibits strong performance with RMSE=0.036. The 97.5\% confidence interval is shown as the shaded region.}
  \vspace{-20pt}
  \label{fig:tnr_tpr}
\end{wrapfigure}

Figure~\ref{fig:tnr_tpr} shows the LOSS TNR to be a very good predictor of LiRA's TPR@FPR=$10^{-3}$ across models and datasets, obtaining a low RMSE of $0.036$ with a simple linear predictor with one parameter. Bootstrapping with 1000 iterations provides narrow confidence intervals, confirming that the linear model maintains a high degree of certainty over the full range of values. This allows practitioners to accurately estimate the LiRA TPR@FPR without reference models. This is particularly useful for many scenarios that require evaluating many candidate models such as hyperparameter tuning, neural architecture search, or comparing training strategies.

We then compare our approach against an alternate metric, (i) LOSS attack AUC, and baselines: (ii) train-test accuracy gap of the target model; and (iii) LT-IQR AUC~\citep{pollock2025free}. LT-IQR is the SOTA for identifying the samples most at risk to LiRA, but is not a measure of overall model vulnerability. 

Table \ref{tab:performance_metrics} reports goodness-of-fit metrics for linear functions fit on our approach and baselines, showing it to outperform the baselines and to be a good estimator of vulnerability to LiRA. 

Aware of the prohibitive cost of SOTA MIA methods, a recent line of research aims to develop low-cost attacks which can be used by model developers to estimate the model-level risk. We also instantiate the SOTA low-cost attack, online RMIA, with two reference models and compare its predictive power to our method. 
As shown in Table~\ref{tab:performance_metrics}, our method outperforms RMIA at estimating LiRA TPR, despite RMIA requiring at least 2 reference models.

LOSS AUC, which measures distributional separability across all loss values, would be more effective if memorization were evenly distributed. The train-test gap, a metric studied in the literature, captures only mean differences between loss distributions. While it performs moderately well, we observe that it cannot detect tail differences caused by the asymmetric non-member distribution.




\begin{table}[t]
\centering
\caption{Performance of reference-free statistics and RMIA as predictors of LiRA TPR@FPR$=10^{-3}$, across the 10 image-classification architectures and 4 datasets of Section~\ref{sec:setup}. Each statistic is mapped to LiRA TPR by a one-parameter least-squares fit anchored at that statistic's no-signal point: the origin for tail statistics, gaps and distances, and chance ($0.5$) for AUC-type statistics. Baselines are ordered by RMSE; our estimator is set off below them. LOSS TNR is the strongest predictor, ahead even of RMIA, which requires training two reference models.}
\label{tab:performance_metrics}
\begin{tabular}{lccc}
\toprule
Metric & $R^2$ & RMSE & MAE \\
\midrule
RMIA (2 reference models) & 0.909 & 0.046 & 0.035 \\
LT-IQR AUC & 0.803 & 0.068 & 0.046 \\
LOSS AUC & 0.802 & 0.068 & 0.049 \\
KS distance & 0.787 & 0.070 & 0.052 \\
Train-test accuracy gap & 0.681 & 0.086 & 0.058 \\
\midrule
\textbf{LOSS TNR} (ours) & \textbf{0.944} & \textbf{0.036} & \textbf{0.025} \\
\bottomrule
\end{tabular}
\end{table}

\paragraph{The relationship holds on unseen architectures and datasets.} The fit shown in  Figure~\ref{fig:tnr_tpr} also holds on unseen setups. It achieves $\mathrm{RMSE}=0.038$ vs. $0.036$ in sample under leave-one-architecture out. We also leave-one-dataset out ($\mathrm{RMSE}=0.064$), which is more challenging as holding out CIFAR-100 forces the fit to extrapolate. The ranking within datasets is unaffected, with $\rho=0.92$--$0.95$. We report the full results in Appendix~\ref{app:heldout}.

\begin{figure}[h]
    \centering
        \centering
        \includegraphics[width=0.36\textwidth]{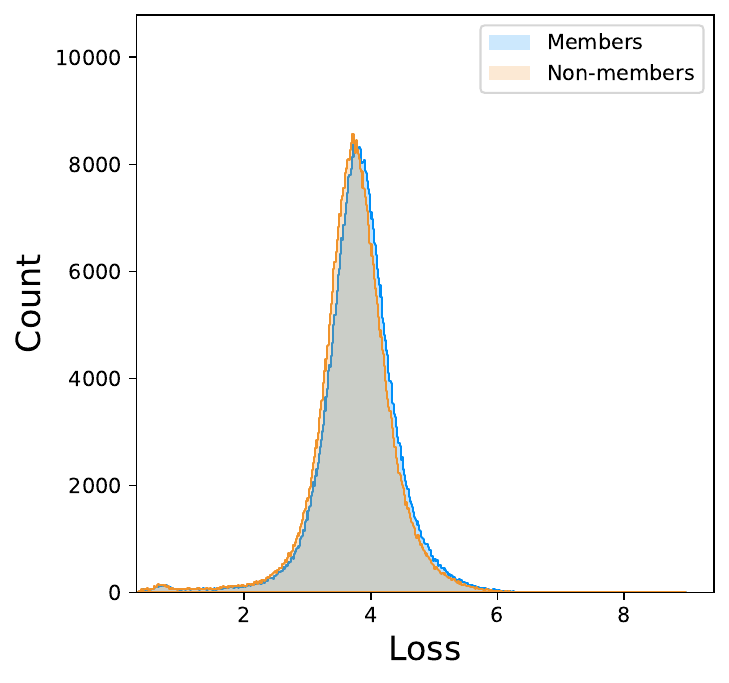}
        \includegraphics[width=0.35\textwidth]{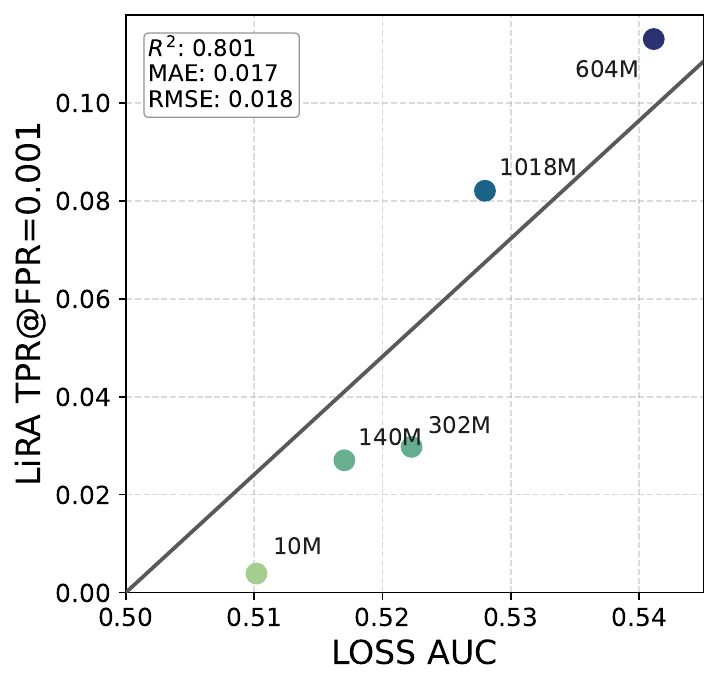}
    \caption{\textbf{$\ell_{\text{target}}$ distributions and predictive relationship for large language models}. Left: $\ell_{{target}}$ distributions for members (orange) and non-members (blue) show high overlap and symmetry, unlike the asymmetric, heavy-tailed distributions observed in image classification models (Figure~\ref{fig:intuition}). Right: Despite symmetric distributions, LOSS AUC reliably predicts LiRA TPR@FPR=0.001 across five GPT-2 models (10M to 1018M parameters), achieving RMSE=0.018. Points are labeled by model size.}
    \label{fig:llms}
\end{figure}

\subsection{LLMs}
\label{sec:llms}

Section \ref{sec:intuition} identifies a second reference-free proxy at the symmetric end of the continuum, where training does not meaningfully collapse $\sigma^2$ at the trained sample and only the residual mean-shift term of Eq.\ref{eq:lira-full} survives.

Recent evidence suggests that LLMs exhibit fundamentally different memorization patterns than traditional models. The size of LLMs, for instance, allows them to achieve excellent generalization while simultaneously memorizing training data~\citep{tirumala2022memorization} and redundancy in large-scale text corpora means that information at $(x,y)$ is typically already represented through near-duplicates elsewhere in training, what~\cite{shilov2024mosaic} describe as \textbf{mosaic memory}. Including a specific $(x, y)$ therefore does not collapse the model's uncertainty at $x$ to the extent it does in image classifiers, where each sample carries roughly unique information. Furthermore, the sequence-level loss is an average over many tokens, most carrying no membership signal, which dilutes whatever signal the relevant tokens do carry, weakening the membership signal~\citep{tao2025token}.

We test this directly. Following the setup proposed in recent work~\citep{hayes2025strongmembershipinferenceattacks}, we evaluate our method on 5 GPT-2 models~\citep{radford2019language} with 10M, 140M, 302M, 604M, and 1018M parameters. We train each model on a $2^{19}$-sample subset of the C4 dataset~\citep{raffel2020exploring} and instantiate online LiRA with $K=127$ reference models. 

Figure \ref{fig:variance-scatter} shows that across all five GPT-2 sizes, $\sigma^2_\text{in}$ and $\sigma^2_\text{out}$ concentrate close to $y=x$, with visibly less member variance collapse than in any image setup. The corresponding loss distributions (Fig. \ref{fig:llms}, left) are near symmetric and identical in contrast to the heavy-tailed image-model distributions. With no heavy tail for members to be missing from, TNR has no signal to pick up. The mean-shift residual remains, and Fig. \ref{fig:llms}(right) shows LOSS AUC to be a strong predictor of LiRA TPR@FPR=$10^{-3}$ across all five GPT-2 sizes, with RMSE 0.018.

We also test this on another family of LLMs, eight Pythia models (14M to 2.8B parameters) trained from scratch on the Pile. Due to the prohibitive cost of running LiRA, we use RMIA as a lower cost proxy for LiRA and find that the same relationship holds. LOSS AUC predicts RMIA at TPR@FPR$=10^{-3}$ with $R^2=0.862$ and $\mathrm{RMSE}=0.024$. We expand on this in Appendix~\ref{app:pythia}.

A practitioner selects a proxy for a given target model using the model's train and test losses. Let $s$ denote the KS distance between the non-member loss distribution and a mean-shifted Gaussian fit to the member losses. Use LOSS AUC if $s\leq 0.5$ and LOSS TNR otherwise. We validate this in Appendix~\ref{app:proxy-selection}.





\section{Conclusion}
\label{sec:conclusion}
We present a framework for estimating model-level vulnerability to SOTA membership inference attacks without reference models. We show that LiRA's per-sample signal decomposes into a variance-ratio term and a residual mean-shift term, with the relative contribution of each determined by how much training collapses model uncertainty at the trained sample. This places models on a continuum, with the shape of the loss distribution itself acting as a reference-free diagnostic for where a given model sits and which loss-based proxy of LiRA TPR is appropriate.

We instantiate the framework with two natural proxies. At the heavy-tailed end of the continuum, where image classifiers sit, the LOSS attack TNR captures the proportion of high-loss outliers absent from the training side, and predicts LiRA TPR@FPR=$10^{-3}$ with RMSE 0.036 across 10 architectures and 4 datasets, outperforming the train-test accuracy gap, LT-IQR, and low-cost reference-model attacks. The method is robust across generalization gaps from near-zero to over 50 percentage points, across FNR threshold choices, and across reference-model attack families: it predicts the TPR of RMIA and Attack R with similar accuracy, confirming that it captures signal shared across this family of MIAs rather than a LiRA-specific artifact. At the symmetric end of the continuum, where LLMs sit, the LOSS attack AUC captures the surviving population-level mean-shift signal, and predicts LiRA TPR with RMSE 0.018 across five GPT-2 sizes from 10M to 1B parameters.

\looseness=-1
While our estimator is empirically calibrated, it is not an arbitrary regression. Both LOSS TNR and LOSS AUC arise as population-level summaries of the same quantities that LiRA thresholds in the two limiting regimes. The observed linear relationship reflects shared dependence rather than incidental correlation.

Where SOTA attacks require training tens to hundreds of reference models, our framework uses only the target model's loss values on its train and test sets. We believe this will substantially help enable privacy risk assessments in practice, in particular during iterative development workflows and scenarios where multiple candidate models require assessment.


\paragraph{Limitations and future work.} While we evaluate three reference-model MIAs (LiRA, RMIA, Attack R), we have not assessed transferability to attacks from fundamentally different families or to broader privacy threats such as reconstruction or inversion. Our experiments span diverse setups across the continuum, but the framework remains an estimator and may not generalize to all scenarios. Our LLM study covers five GPT-2 models with LiRA and eight Pythia models with RMIA, spanning 10M to 2.8B parameters and two corpora. Scaling to larger models and more datasets is a natural next step. The framework is also proxy-agnostic, and other cheap MIA signals such as Min-K\%++~\citep{zhang2024min} are natural candidates for additional instances.



\begin{ack}
We thank Jamie Hayes for his support on the experiments related to LLMs, and for providing valuable feedback. 

Additional revenues related to this work: Stipend and compute from MATS Research; Stipend and compute from Anthropic Fellows Program.

\end{ack}




\bibliography{bibliography}
\bibliographystyle{icml2026}

\appendix

\section{Related Work}
\label{sec:related_work}
\paragraph{Membership inference attacks.} MIAs aim to infer whether a data sample was used to train a given target model. ~\citet{homer2008resolving} originally introduced membership inference attacks (MIA) to identify the presence of an individual's genome data among aggregate summary statistics.~\citet{shokri2017membership} proposed the first MIA on machine learning models and demonstrated their efficacy in the black-box setting. MIAs have since become the de-facto standard to empirically measure the privacy risk posed by a model and providing an upper bound on the risk of more severe empirical attacks such as attribute inference or model inversion~\citep{10.1145/2810103.2813677}. 

Simple MIAs compare metrics like loss~\citep{yeom2018privacy} or confidence~\citep{song2019privacy} against thresholds, while others compare samples against neighboring samples~\citep{mattern2023membershipinferenceattackslanguage}. While these attacks can obtain high AUC values, they have been shown to be unable to confidently identify members as they struggle to distinguish between low-loss samples that have been memorized and those that are easy to predict. Difficulty-calibrated attacks~\citep{watson2021importance, song2020systematicevaluationprivacyrisks, he2024difficultycalibrationneedpractical} aim to address this but still lack high confidence.

State-of-the-art attacks rely on reference models~\citep{shokri2017membership} to calibrate the attack with respect to sample difficulty. The Likelihood Ratio Attack (LiRA)~\citep{carlini2022membership} estimates loss distributions for IN and OUT models and performs a likelihood ratio test to determine membership of a given sample. Other methods such as Attack R use only OUT reference models~\citep{ye2022enhanced}, while others employ knowledge distillation to train reference models~\citep{LiuTrajectoryMIA2022,li2024seqmia}.

\paragraph{Low-cost attacks and free identification of vulnerable samples.}
\looseness-1
Reference model-based attacks depend heavily on the number and quality of reference models. On CIFAR-10, LiRA TPR@$0.01\%$ drops from $2.83\%$ with 254 reference models to $0.76\%$ with 4~\citep{zarifzadeh2023low}. Performance declines further with fewer models or smaller or mismatched architectures. In practice, dozens to hundreds of identical reference models are often needed for high accuracy, making these attacks difficult to scale to iterative workflows and large-scale models such as LLMs~\citep{carlini2022membership}.

Recent work has explored ways to estimate model vulnerability to LiRA at lower computational cost or even for free. The state-of-the-art low-cost attack, RMIA~\citep{zarifzadeh2023low}, combines reference models with population samples to construct an attack they show to match LiRA performance with as few as 2 reference models. Quantile regression-based attacks~\citep{bertran2023scalable} use a single regression model to match LiRA's TPR, but their effectiveness is limited to scenarios with only one reference model. 

Another direction of research focuses on free identification of the samples that are most vulnerable to SOTA reference model-based MIAs. Predictors such as sample loss, confidence, and training loss dynamics have been shown to successfully identify highly vulnerable samples~\citep{pollock2025free, leemann2024my} for free. In particular, LT-IQR has been shown to achieve high precision at identifying samples vulnerable to LiRA~\citep{pollock2025free}. These methods, however, do not readily provide model-level risk scores as their outputs are inherently relative.

\paragraph{Factors influencing membership inference vulnerability.}
Previous work has explored the connection between a model's training and its vulnerability to MIAs.~\citet{yeom2018privacy} proposed overfitting as a sufficient but not necessary condition for MIA success, whilst other works demonstrate a correlation between the train-test gap (a measure of overfitting) and MIA success~\citep{yeom2018privacy, carlini2022membership}. Other work has studied the impact of dataset properties on model vulnerability: models trained on smaller datasets are more vulnerable~\citep{tobaben2025impactdatasetpropertiesmembership, chen2020gan, németh2025privacyaccuracyimplicationsmodel}, and, within a dataset, classes with fewer samples contain more vulnerable examples~\citep{kulynych2022disparate, chang2021privacy}.

\label{sec:appendix}

\section{Why each proxy is appropriate at its end of the continuum}
\label{app:proxy-comparison}

\begin{figure}[H]
    \centering
    \includegraphics[width=\linewidth]{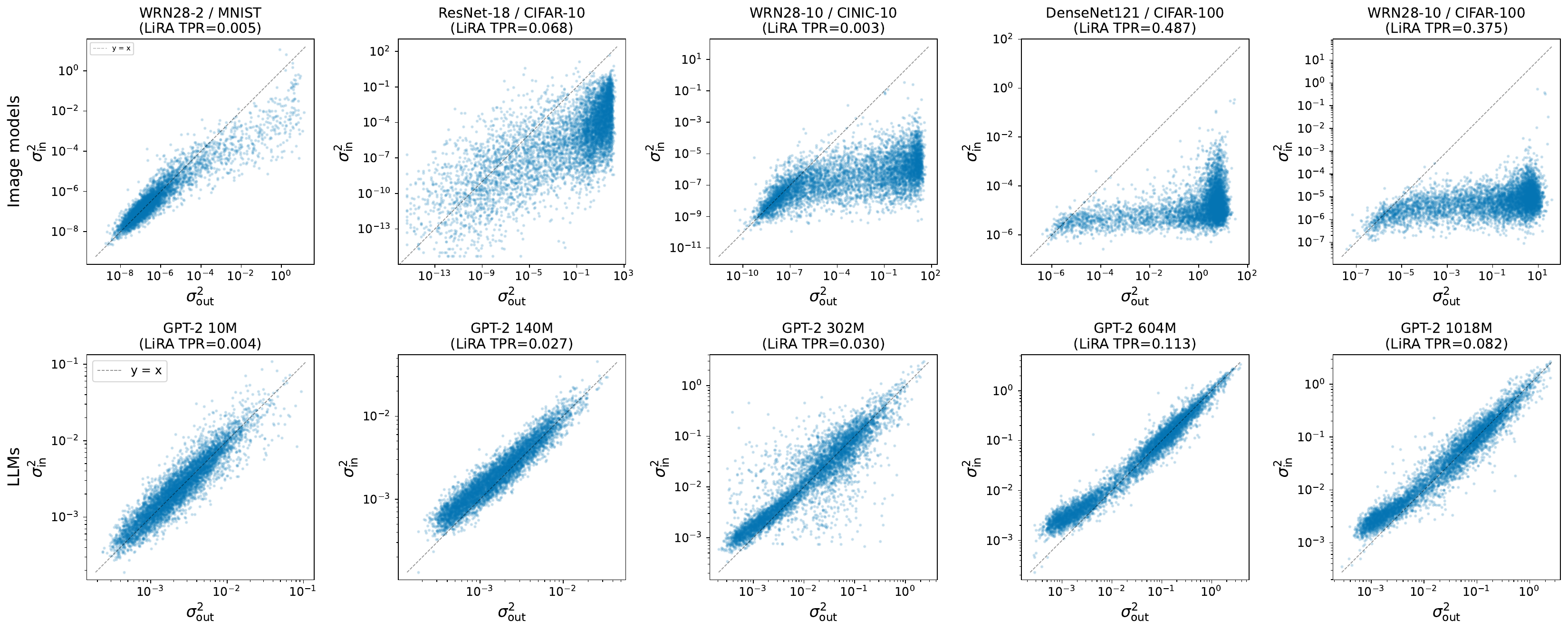}
    \caption{\textbf{Training collapses per-sample uncertainty in image
models but not in LLMs.} Each panel plots $\sigma^2_\text{in}$ against
$\sigma^2_\text{out}$ per sample. \textbf{Top row (image models):}
the members are scattered below $y = x$, indicating that training has collapsed
the predictive variance at trained samples. \textbf{Bottom row (GPT-2):}
the points sit close to $y = x$ at all scales. $\sigma^2_\text{out}$ itself
spans several orders of magnitude indicating that training on a specific sample simply does
not meaningfully reduce it rather than it being uniformly low. This places image models near the
strong-collapse end of the continuum (Section~\ref{sec:intuition}) and
LLMs near the weak-collapse end.}
\label{fig:variance-scatter}
\end{figure}

Section~\ref{sec:gp-analysis} places models on a continuum between two limits, with the loss-distribution shape as the reference-free diagnostic for where a model sits. Figure~\ref{fig:variance-scatter} confirms this directly: image classifiers exhibit substantial member variance collapse and sit near the strong-collapse end of the continuum, while LLMs exhibit much less and sit near the weak-collapse end.

We now formalize the corresponding proxy selection. The LiRA decomposition (Eq.~\eqref{eq:lira-full}) splits the per-sample score into a variance-ratio term and a residual term, with one or the other dominating at each end of the continuum. Whichever term dominates determines a population-level parameter that drives LiRA TPR. In the weak-collapse limit, LOSS TNR and LOSS AUC are functions of the same parameter, and the question is which estimates it more efficiently from a finite sample. In the strong-collapse limit, LOSS TNR tracks the quantity exploited by LiRA directly, while LOSS AUC also absorbs a bulk term that LiRA largely ignores.

We work at the population level. $\mathcal{L}_\text{mem}$ and $\mathcal{L}_\text{non}$ denote the population distributions of $\ell_\text{target}$ on $\mathcal{D}_\text{train}$ and $\mathcal{D}_\text{test}$ respectively. The two proxies are
\begin{equation}
\label{eq:proxies}
\text{AUC} = \Pr_{X \sim \mathcal{L}_\text{mem},\, Y \sim \mathcal{L}_\text{non}}(X < Y), \qquad
\text{TNR}(\beta) = \Pr_{\ell \sim \mathcal{L}_\text{non}}(\ell > \tau_\beta),
\end{equation}
with $\tau_\beta$ chosen so that $\Pr_{\ell \sim \mathcal{L}_\text{mem}}(\ell > \tau_\beta) = \beta$ (the LOSS attack false negative rate). Following the main text, we set this FNR equal to the LiRA FPR at which we predict TPR, so $\beta$ denotes both. Both are computed from $\ell_\text{target}$ on $\mathcal{D}_\text{train}$ and $\mathcal{D}_\text{test}$ alone and require no reference models. We relate each proxy to LiRA TPR at fixed FPR $\beta$, in each of the two regimes derived in Section~\ref{sec:gp-analysis}, and close with the unifying principle.

\subsection{Weak-collapse limit}
\label{app:weak-collapse}

In the weak-collapse limit of Eq.~\eqref{eq:lira-regime2}, per-sample variance is unaffected by training and the population-level loss distributions are unimodal and near-symmetric. We model them as $\mathcal{L}_\text{mem} \sim \mathcal{N}(\bar\mu_\text{mem}, \bar\sigma^2)$ and $\mathcal{L}_\text{non} \sim \mathcal{N}(\bar\mu_\text{non}, \bar\sigma^2)$ with shared variance, and write
\begin{equation}
\label{eq:d-definition}
d = \frac{\bar\mu_\text{non} - \bar\mu_\text{mem}}{\bar\sigma}
\end{equation}
for the standardized mean separation. (Bars distinguish these population-level moments from the per-sample moments below)

The FNR constraint $\Pr_{\ell \sim \mathcal{L}_\text{mem}}(\ell > \tau_\beta) = \beta$ gives $\tau_\beta = \bar\mu_\text{mem} + z_{1-\beta}\bar\sigma$, where $\Phi$ is the standard normal CDF and $z_q = \Phi^{-1}(q)$. Substituting into the TNR expression, splitting the fraction, and applying $1 - \Phi(x) = \Phi(-x)$:
\begin{equation}
\label{eq:tnr-symmetric}
\text{TNR}(\beta) = 1 - \Phi\!\left(\frac{\tau_\beta - \bar\mu_\text{non}}{\bar\sigma}\right) = 1 - \Phi(z_{1-\beta} - d) = \Phi(d - z_{1-\beta}).
\end{equation}

Variances add for independent random variables, so $Y - X \sim \mathcal{N}(\bar\mu_\text{non} - \bar\mu_\text{mem}, 2\bar\sigma^2)$. Then
\begin{equation}
\label{eq:auc-symmetric}
\text{AUC} = \Pr(Y - X > 0) = \Phi(d/\sqrt{2}).
\end{equation}
While LOSS thresholds a population-level statistic, LiRA is a per-sample
attack, so the two operate at different noise scales. For each sample $x$,
let $\mu_\text{mem}(x)$ and $\mu_\text{non}(x)$ denote the mean of its loss
over training runs that include and exclude $x$, and $\sigma^2(x)$ its
variance over training runs, which is unchanged by training in this limit.
Under the GP, these play the role of the predictive moments of
Section~\ref{sec:gp-analysis}. We decompose the population variance of the
loss into a within-sample and a between-sample component:
\begin{equation}
\label{eq:var-decomp}
  \bar\sigma^2 = \sigma_w^2 + \sigma_b^2, \qquad
  \sigma_w^2 := \mathbb{E}_x\big[\sigma^2(x)\big], \qquad
  \sigma_b^2 := \text{Var}_x\big[\mu_\text{mem}(x)\big],
\end{equation}
where $\sigma_b^2$ captures variation in sample difficulty. For a
homogeneous shift $\delta = \mu_\text{non}(x) - \mu_\text{mem}(x)$,
$\sigma_b^2$ is the same for members and non-members. LiRA standardizes each
sample's loss against its own reference distribution, so its score is a
monotone function of $(\ell - \mu_\text{non}(x))/\sigma_w$. This is
distributed as $\mathcal{N}(-d_\text{LiRA}, 1)$ on members and
$\mathcal{N}(0, 1)$ on non-members, with $d_\text{LiRA} = \delta/\sigma_w$.
LOSS instead sees the population separation
\begin{equation}
\label{eq:rho}
  d = \frac{\delta}{\bar\sigma} = \rho\, d_\text{LiRA}, \qquad
  \rho := \Big(1 + \sigma_b^2 / \sigma_w^2\Big)^{-1/2} \le 1.
\end{equation}
By the same Gaussian-tail computation as Eq.~\eqref{eq:tnr-symmetric}, the
three quantities are
\begin{equation}
\label{eq:weak-three}
  \text{TPR}_\text{LiRA}(\beta) \approx \Phi(d_\text{LiRA} - z_{1-\beta}), \quad
  \text{TNR}(\beta) = \Phi(\rho\, d_\text{LiRA} - z_{1-\beta}), \quad
  \text{AUC} = \Phi\big(\rho\, d_\text{LiRA} / \sqrt{2}\big).
\end{equation}

\paragraph{Signal-to-noise.} The proxies separate in finite samples. Both
see only $d = \rho\, d_\text{LiRA}$, and any proxy noise propagates through
the calibration into the predicted LiRA TPR. What matters is how much noise
each proxy carries relative to how strongly it moves with $d$.

TNR is a binomial proportion. The empirical estimator counts
$\mathcal{D}_\text{test}$ samples above $\tau_\beta$, with standard error
\begin{equation}
\label{eq:se-tnr}
\text{SE}(\widehat{\text{TNR}}) = \sqrt{\frac{\text{TNR}(1 - \text{TNR})}{|\mathcal{D}_\text{test}|}} \approx \sqrt{\frac{\text{TNR}}{|\mathcal{D}_\text{test}|}}
\end{equation}
when TNR is small. At the low FNR of interest ($\beta = 10^{-3}$) and
$d = 0.5$, Eq.~\eqref{eq:tnr-symmetric} gives
$\text{TNR}(10^{-3}) \approx 5 \times 10^{-3}$. With
$|\mathcal{D}_\text{test}| = 10^4$ the standard error is roughly
$7 \times 10^{-4}$. The slope in $d$ is only
$\varphi(d - z_{1-\beta}) \approx 0.014$, so the noise on $d$ is about $0.05$.

As AUC is a U-statistic over pairs, every (member, non-member) pair contributes, but pairs sharing a sample are correlated, so
\begin{equation}
\label{eq:se-auc}
\text{SE}(\widehat{\text{AUC}}) \sim \frac{1}{\sqrt{\min(|\mathcal{D}_\text{train}|, |\mathcal{D}_\text{test}|)}}
\end{equation}
is independent of $\beta$ \citep{hanley1982meaning}. For
$|\mathcal{D}_\text{train}| = |\mathcal{D}_\text{test}| = 10^4$ and $d = 0.5$,
AUC is around $0.638$ and the standard error is roughly $4 \times 10^{-3}$.
The slope in $d$ is $\varphi(d/\sqrt{2})/\sqrt{2} \approx 0.27$, so the noise
on $d$ is about $0.015$.

The two estimators have different effective sample sizes. TNR uses only the samples that fall above $\tau_\beta$ (a tiny fraction of the test set at low), and sits in the tail where it barely moves with $d$. AUC, on the other hand, uses every pair. Its noise on $d$ is about three times smaller making it the more reliable predictor of LiRA TPR in this limit.

\subsection{Strong-collapse limit}
\label{app:strong-collapse}

In the strong-collapse limit of Eq.~\eqref{eq:lira-regime1}, training collapses the per-sample variance on members. The non-member distribution develops a heavy tail, and the member distribution loses part of that tail to training-induced migration into the head. The two distributions now differ in shape rather than location, so a mean-shift model no longer applies.

We instead use the structure of the train/test split. Members and non-members are drawn i.i.d., so $\mathcal{L}_\text{non}$ is also the distribution each member's loss would have followed had it not been trained on. We denote this counterfactual loss as  $\ell_\text{out}(x)$. As before, $\tau_\beta$ is the $(1-\beta)$-quantile of $\mathcal{L}_\text{mem}$. We
assume training does not push members up across the threshold: if
$\ell_\text{out}(x) \le \tau_\beta$, then $\ell_\text{target}(x) \le \tau_\beta$.
We call this (A1). The migrated fraction
\begin{equation}
m(\tau) = \Pr_{x \in \mathcal{D}_\text{train}}\big(\ell_\text{out}(x) > \tau,\ \ell_\text{target}(x) \le \tau\big)
\end{equation}
is the fraction of members that would have sat above $\tau$ unseen but sit
below it after training.

\paragraph{TNR.} Since non-members are drawn from the same distribution, a fraction $\text{TNR}(\beta)$ of members have $\ell_\text{out} > \tau_\beta$. By construction, a fraction $\beta$ of members are still above $\tau_\beta$ after training, and by (A1) these belong to the first group. The rest migrated:
\begin{equation}
\label{eq:tnr-identity}
\text{TNR}(\beta) = m(\tau_\beta) + \beta.
\end{equation}
TNR is a direct estimate of the migrated fraction. This holds at any $\beta$ and makes no assumption on the shape of either distribution.

\paragraph{LiRA.} Migrated members have high $\bar\ell_\text{out}$ and low $\ell_\text{target}$. Both terms of Eq.~\eqref{eq:lira-regime1} are large for them, and they are the samples LiRA flags at low FPR
(Figure~\ref{fig:intuition}). Non-migrated
members shift too little relative to $\sigma_\text{out}(x)$ to clear the
threshold. Let $\kappa(\beta, K)$ be the fraction of migrated members that LiRA with $K$ reference models identifies at FPR $\beta$. Then
\begin{equation}
\label{eq:lira-tnr}
\text{TPR}_\text{LiRA}(\beta) \approx \kappa(\beta, K)\, m(\tau_\beta) = \kappa(\beta, K)\big(\text{TNR}(\beta) - \beta\big).
\end{equation}
This predicts a linear fit with intercept close to zero, and a slope that
grows with attack strength. Appendix~\ref{app:vary_k} confirms that the slope increases with $K$. That
$\kappa$ is roughly constant across architectures and datasets for fixed
$K$ is an empirical finding, not a consequence of the model.

\paragraph{AUC.} Let $X \sim \mathcal{L}_\text{mem}$ and
$Y \sim \mathcal{L}_\text{non}$, and condition on whether $Y$ lies above
$\tau_\beta$. A non-member above $\tau_\beta$ exceeds all but a fraction
$\beta$ of members, so
\begin{equation}
\label{eq:auc-decomp}
\text{AUC} \approx \text{TNR}(\beta) + \big(1 - \text{TNR}(\beta)\big)\, a_\text{bulk},
\qquad a_\text{bulk} = \Pr(X < Y \mid Y \le \tau_\beta).
\end{equation}
AUC depends on the migrated fraction through TNR, but also on
$a_\text{bulk}$, the ordering of members and non-members within the head.

Comparing Eqs.~\eqref{eq:tnr-identity}, \eqref{eq:lira-tnr},
and~\eqref{eq:auc-decomp}: TNR and LiRA TPR depend on the same migrated
fraction. AUC depends on it too, plus a bulk term.

\paragraph{Why TNR is preferred.} $a_\text{bulk}$ reflects how much training lowers the loss of typical samples. This varies with architecture, augmentation, and training length, independently of tail migration. LiRA at FPR $= 10^{-3}$ largely ignores it, since small shifts on easy samples do not clear a strict threshold. For AUC it is noise: it adds variation across models that LiRA TPR does not share. TNR excludes it. This is consistent with Table~\ref{tab:performance_metrics}, where the train-test loss gap (which only measures the shift in mean loss) is a weak predictor compared to LOSS TNR.

\paragraph{Unifying view.} Eq.~\eqref{eq:auc-decomp} holds in both regimes. In the weak-collapse limit,
training shifts every member's loss by a small amount, so only members
already near $\tau_\beta$ cross it. The migrated fraction $m$ is small, and TNR stays close to $\beta$, where it is dominated by noise. The difference between members and non-members then sits almost entirely in
$a_\text{bulk}$, which is the mean shift of Appendix~\ref{app:weak-collapse}. In the strong-collapse limit, training moves a substantial fraction of members out of the tail, and TNR carries the signal. The continuum of Section~\ref{sec:gp-analysis} is a continuum in how much of the member-non-member difference sits in the tail term of Eq.~\eqref{eq:auc-decomp} versus the bulk term.

\section{Selecting a proxy in practice}
\label{app:proxy-selection}

In Section~\ref{sec:method}, we apply different proxies to different model families and elaborate on why each proxy is more appropriate in that regime. A practitioner evaluating a given model must decide which regime their model is in, and consequently which proxy is most suitable. A practitioner has access to the target model and its losses on $\mathcal{D}_\text{train}$ and $\mathcal{D}_\text{test}$, but no ground-truth LiRA TPR to validate against. Proxy selection must therefore rely on diagnostics computed directly from these loss distributions. 

We test whether the non-member loss distribution is well-approximated by a mean-shifted version of the member distribution. Concretely, we fit a constrained Gaussian model $N(\mu + \Delta, \sigma_\text{mem})$ to the non-member losses, where $\mu$ and $\sigma_\text{mem}$ are estimated from $\mathcal{L}_\text{mem}$ and $\Delta$ is a fitted shift. We then compute the Kolmogorov--Smirnov (KS) statistic $s$ between the empirical CDF of $\mathcal{L}_\text{non}$ and this constrained Gaussian. A small KS value indicates that the mean-shift assumption holds (weak-collapse regime), and LOSS AUC is appropriate. A large KS value indicates a systematic deviation from this model e.g., heavy tails, skew.

\paragraph{The Decision Rule.} Use LOSS AUC if $s \leq \tau$ and LOSS TNR otherwise, with $\tau \approx 0.5$. The two regimes separate cleanly on our models: the five GPT-2 models have $\mathrm{KS} \in [0.047, 0.068]$ and all 37 image classifiers have $\mathrm{KS} \in [0.35, 0.89]$, so any threshold within this gap recovers the proxy selections from Section~\ref{sec:method}.

To validate the rule in the intermediate range where the choice is ambiguous, we also add 41 RelaxLoss-defended models ($\mathrm{KS} \in [0.07, 0.81]$) and 626 intermediate training checkpoints ($\mathrm{KS} \in [0.012, 0.89]$), with LiRA instantiated against each. $s$ rises smoothly with training accuracy, empirically confirming the continuum of Section~\ref{sec:intuition}. 

We report the held-out $\mathrm{RMSE}$ on the 83 independent configurations (37 image classifier, 5 GPT-2 models, 41 RelaxLoss models). At the error-minimising threshold $\tau^\star = 0.569$:

\begin{itemize}
    \item Our rule: $0.060$
    \item Always TNR: $0.074$
    \item Always AUC: $0.078$
    \item Per-model oracle: $0.052$
\end{itemize}

The rule outperforms both single-proxy baselines for any threshold in the interval $\tau \in [0.066, 0.587]$, making $\tau \approx 0.5$ a safe recommendation. For borderline models, computing both proxies on a small calibration set and keeping the lower-error one closes most of the remaining gap at negligible cost.

\section{Metric performance along the no-collapse--collapse continuum}\label{app:relaxloss}

\begin{figure}[h]
    \centering
    \includegraphics[width=0.45\linewidth]{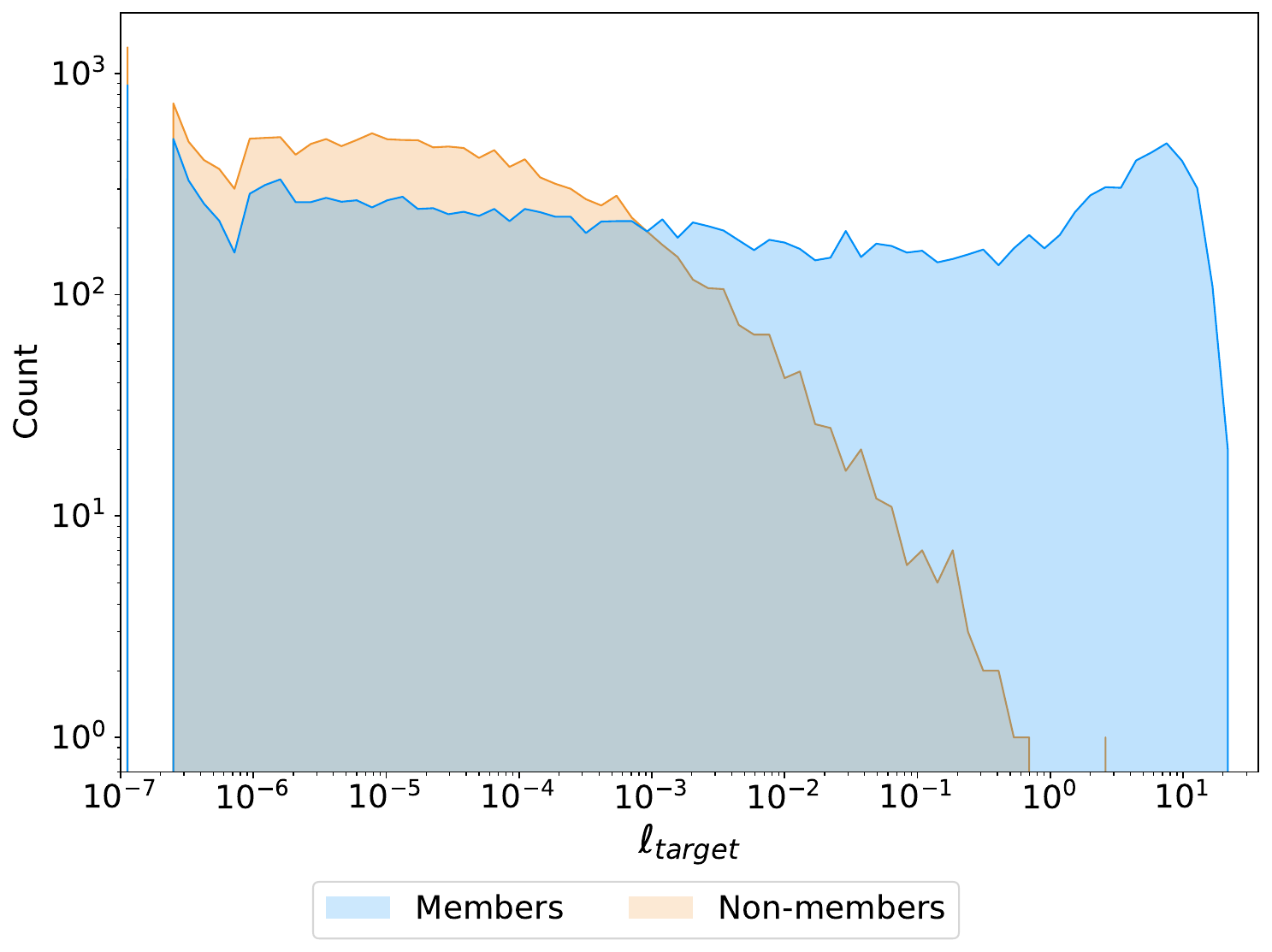}
    \includegraphics[width=0.45\linewidth]{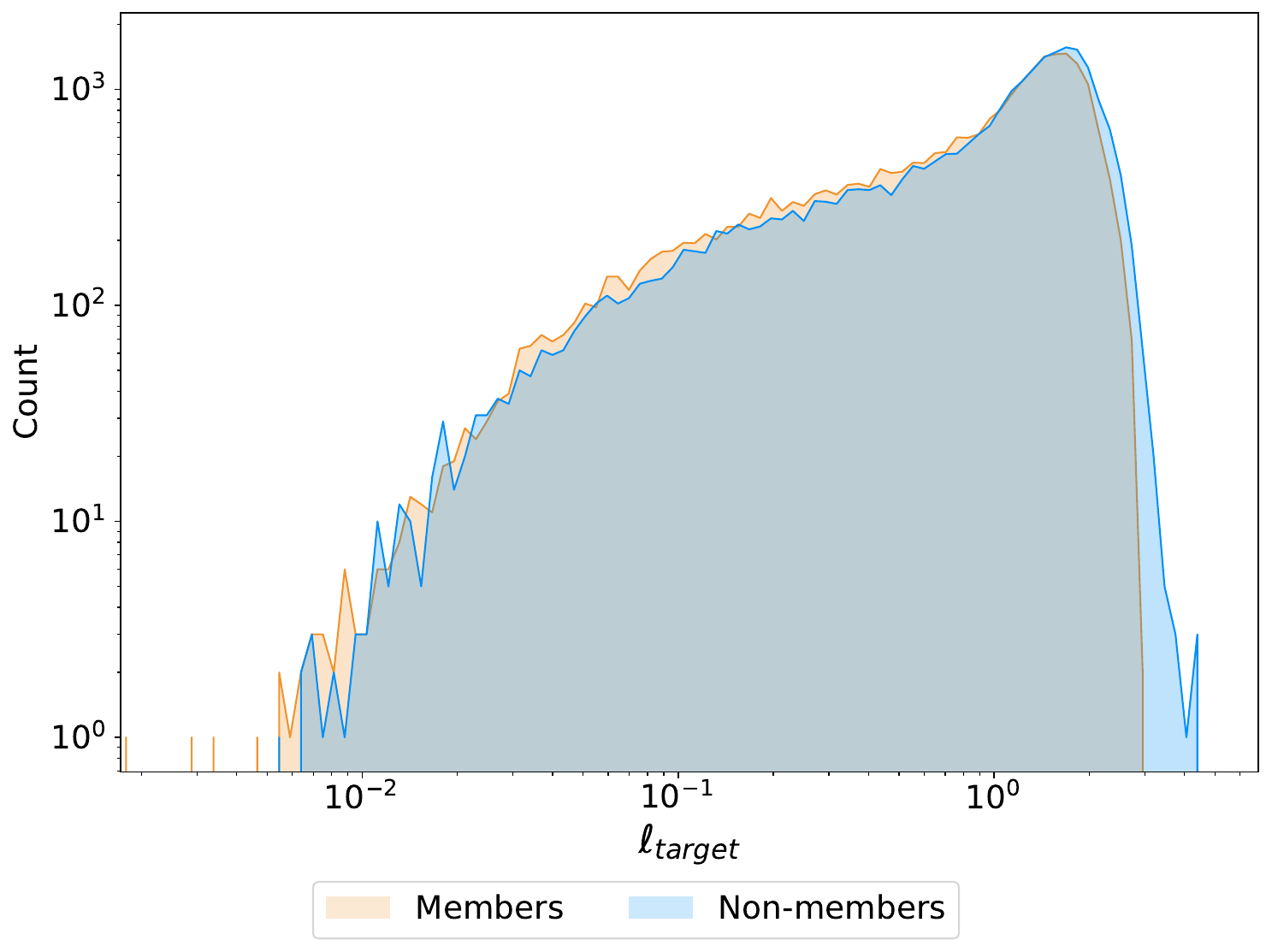}
    \caption{Member and non-member loss distributions for VGG-16 trained on CIFAR-10 with (a) no defense, $s(\mathcal{L}_\text{mem}, \mathcal{L}_\text{non}) = 0.87$, and (b) RelaxLoss with $\alpha=1.0$, $s(\mathcal{L}_\text{mem}, \mathcal{L}_\text{non}) = 0.14$.}
    \label{fig:shortfall_distrib}
\end{figure}

We study how the predictive performance of LOSS TNR and LOSS AUC varies as the member and non-member loss distributions of the target model become more similar. To simulate this, we train target models using RelaxLoss~\citep{chenrelaxloss}, a privacy defense that explicitly encourages the member and non-member loss distributions to converge during training, across 5 architectures (ResNet-20, DenseNet-121, WRN-28-2, VGG-16, VGG-11) and 2 datasets (CIFAR-10, CIFAR-100). We vary the defense strength $\alpha \in \{0, 0.25, 0.5, 0.75, 1.0\}$ and instantiate LiRA against each resulting target model.

Since the same $\alpha$ value has differing effects across datasets, we use the KS value described in~\ref{app:proxy-selection} to measure the distance between the member and non-member loss distributions. Specifically, we fit a constrained Gaussian $N(\mu + \Delta, \sigma_{\text{mem}})$ to the non-member losses. $\mu$ and $\sigma_{\text{mem}}$ are estimated from $\mathcal{L}_{\text{mem}}$ and $\Delta=E[\mathcal{L}_{\text{non}}] - \mu$ is a fitted shift. We use the KS statistic between the empirical CDF of $\mathcal{L}_{\text{non}}$ and this Gaussian directly as a distance measure,
\begin{equation*}
    s(\mathcal{L}_\text{mem}, \mathcal{L}_\text{non}) = \sup_{x \in \mathbb{R}} \left| \hat{F}_\text{non}(x) - \Phi\!\left(\frac{x - (\mu + \Delta)}{\sigma_\text{mem}} \right) \right|,
\end{equation*}

where $\hat F_{\text{non}}$ is the empirical CDF of $\mathcal{L}_{\text{non}}$ and $\Phi$ is the standard normal CDF. A low value of $s$ indicates highly similar distributions (weak-collapse limit), while higher values indicate distributions with different shapes (strong-collapse limit).

\begin{figure}
    \centering
    \includegraphics[width=0.7\linewidth]{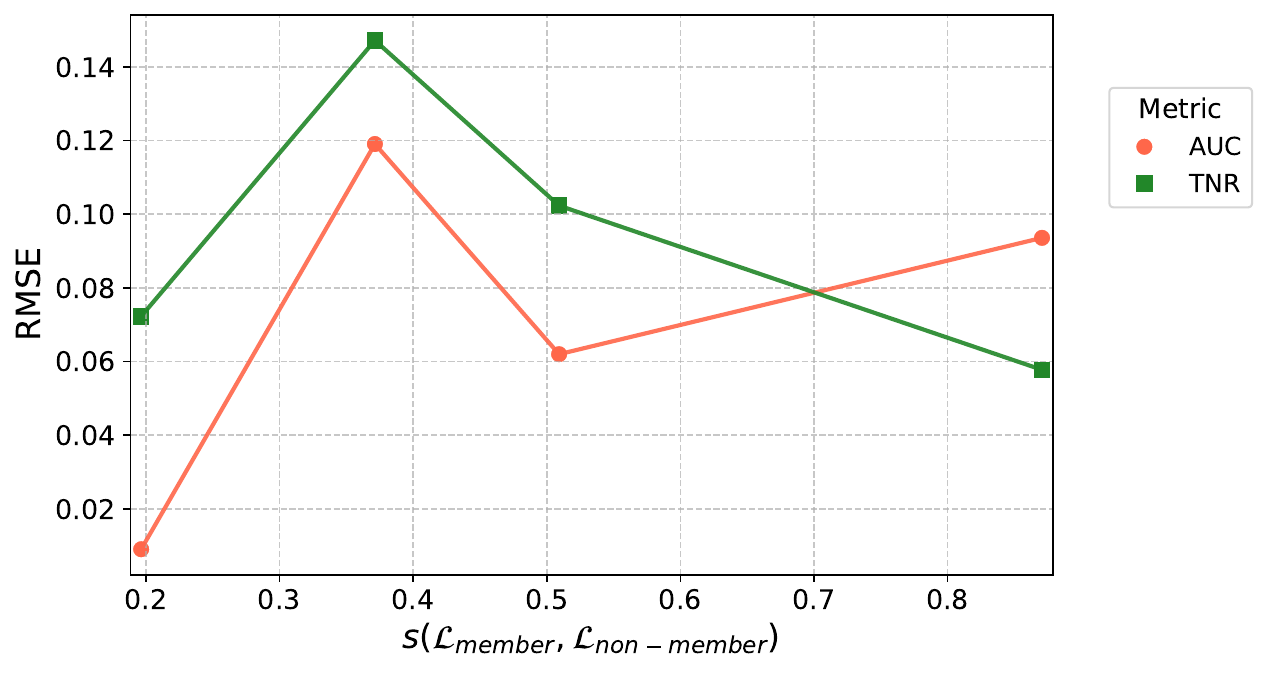}
    \caption{RMSE for predicting LiRA TPR@FPR=$10^{-2}$ with LOSS AUC and LOSS TNR across different member--non-member training distributions. Higher $s$ indicates that the two distributions are more different (strong-collapse limit), and lower values indicate that they are more similar (low-collapse limit).}
    \label{fig:shortfall}
\end{figure}

Figure~\ref{fig:shortfall} shows the RMSE of LOSS TNR and LOSS AUC as predictors of LiRA TPR@FPR=$10^{-2}$ across this range. When the non-member distribution is substantially heavier-tailed than the member distribution (high $s$), LOSS TNR is the stronger predictor. As the two distributions converge, LOSS AUC takes over.

\section{Loss distributions across all setups}
\label{app:dists}

\raggedbottom
\begin{figure}[H]
    \centering
    \includegraphics[width=\linewidth]{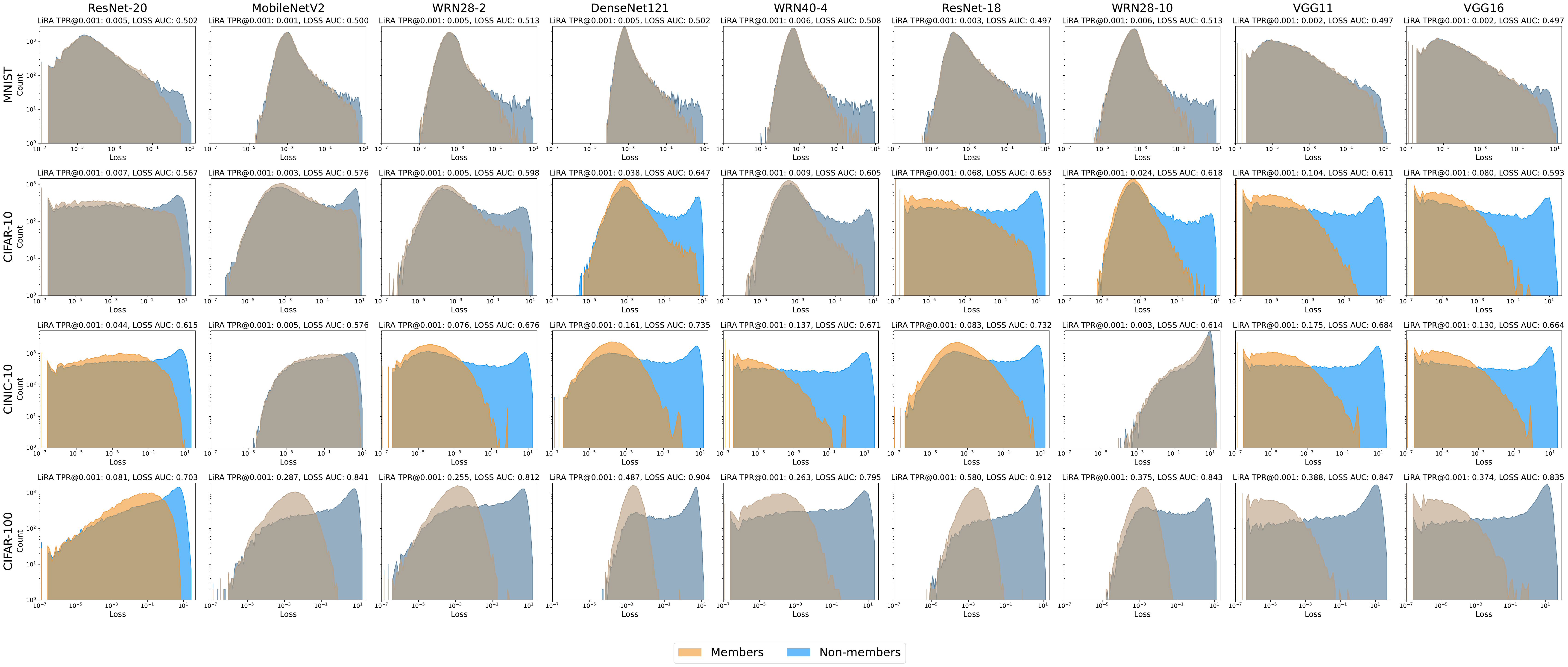}
    \caption{Histograms showing $\ell_{target}$ distributions for training set members (orange) and non-members (blue) in log-log scale across all setups. The distributions demonstrate clear separation between members and non-members, with members typically having lower loss values.}
    \label{fig:member_non_member}
\end{figure}

\begin{figure}[H]
    \centering
    \includegraphics[width=\linewidth]{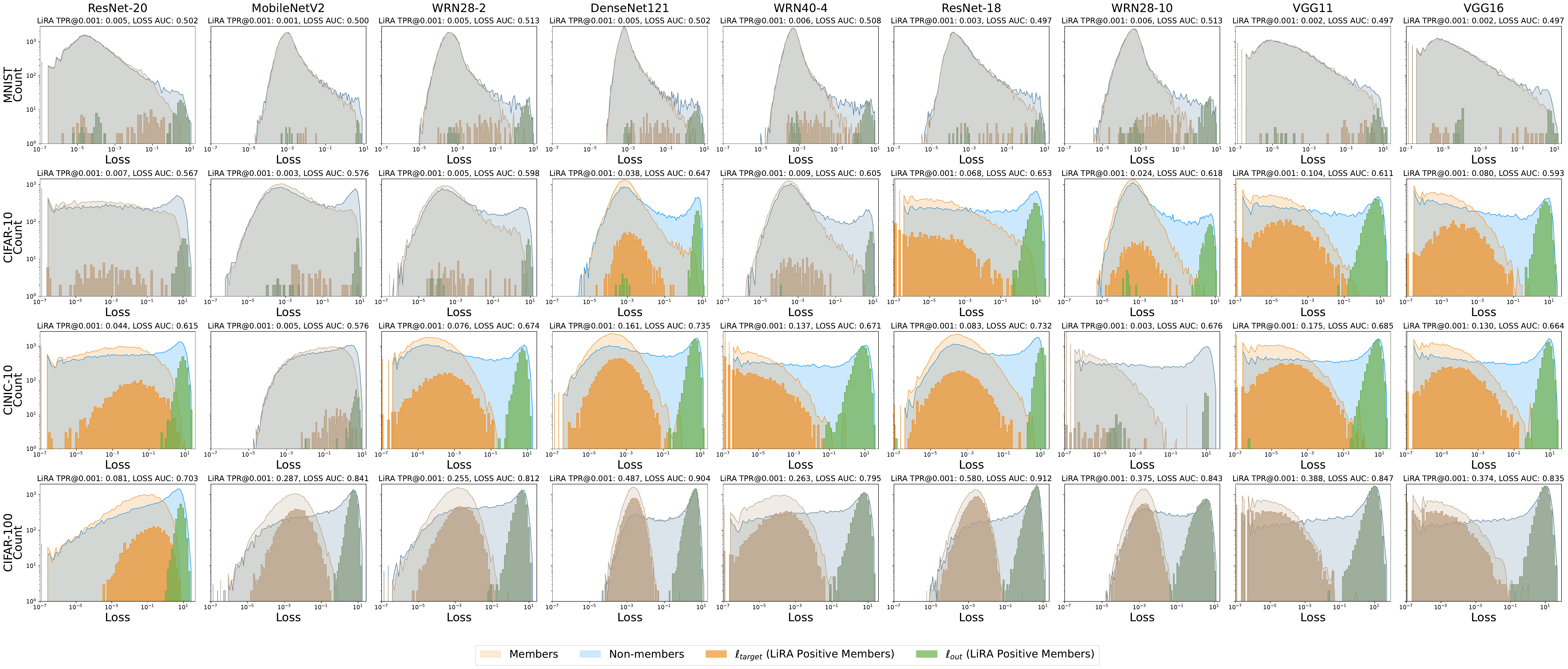}
     \caption{$\ell_{target}$ distributions for training set members and non-members across all setups. Orange histogram shows member losses, blue shows non-member losses, with density curves overlaid. Green bars indicate the OUT model mean ($\ell_{out}$) for points identified by LiRA at FPR=0.001.
    }
    \label{fig:member_non_member_top_lira}
\end{figure}


\section{Varying FPRs}
\label{app:metrics_fprs}
\begin{figure}[H]
    \centering
    \includegraphics[width=\linewidth]{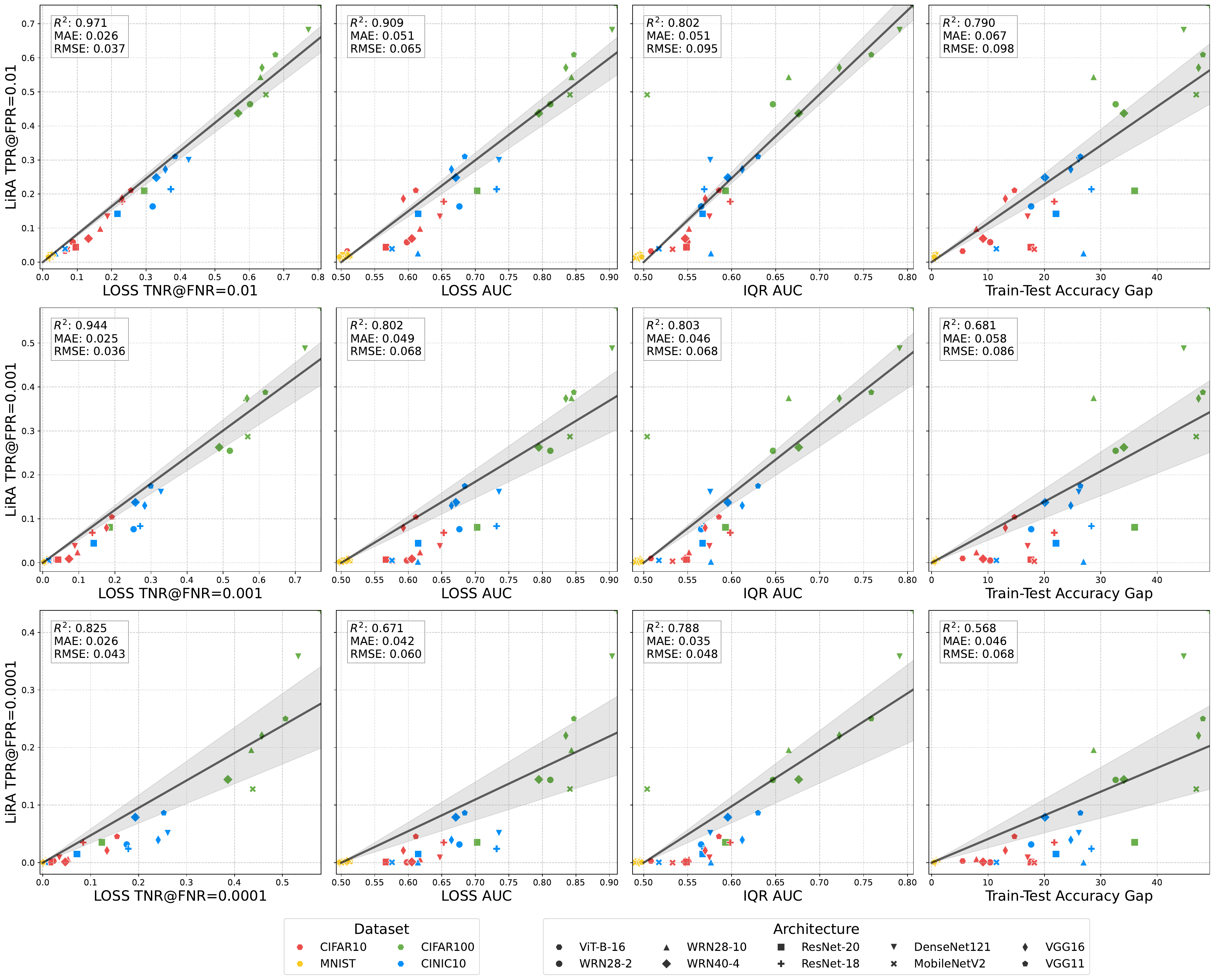}
    \caption{Comparison of different metrics for predicting LiRA TPR at FPRs of 0.01, 0.001, 0.0001. Each panel shows the relationship between a predictor metric (x-axis) and LiRA TPR (y-axis) across all model-dataset combinations. Linear fits (solid lines) with 97.5\% confidence intervals (shaded regions) show goodness-of-fit.}
    \label{fig:varying_fprs}
\end{figure}

\section{Sensitivity to selection of LOSS FNR threshold}
\label{app:sensitivity}
We select FPR=FNR across the paper, as it is a natural choice. We here study the sensitivity of LOSS TNR's predictive capability to the choice of FNR threshold. Following the experimental setup described in Section \ref{sec:setup}, we evaluate how well LOSS TNR predicts LiRA's TPR across four different false positive rate thresholds: 0.1, 0.01, 0.001, and 0.0001. For each LiRA FPR setting, we fit a linear function to predict the corresponding LiRA TPR from LOSS TNR values computed at varying FNR thresholds, and measure prediction quality using the RMSE.

\begin{figure}[H]
    \centering
    \includegraphics[width=\linewidth]{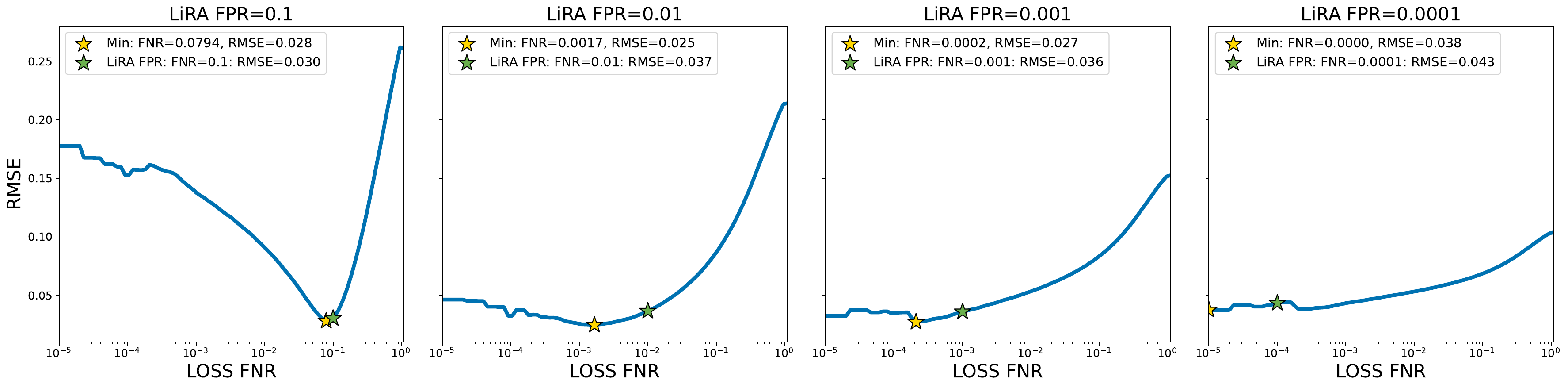}
    \caption{{RMSE of LOSS TNR@different FNR values as a predictor of LiRA TPR at a fixed FPR. The yellow star indicates the FNR that achieves the lowest RMSE, while the green star shows performance when setting the FNR to the LiRA FPR. Across all settings, the RMSE difference between the optimal selection and FNR=FPR is minimal.}}
    \label{fig:vary_fnr}
\end{figure}

Figure \ref{fig:vary_fnr} shows that RMSE remains relatively stable across a broad range of FNR values, demonstrating that LOSS TNR's predictive power is robust to the specific threshold selection. Notably, setting the FNR equal to the target LiRA FPR (green stars) achieves a similar performance compared to the FNR that minimizes RMSE (yellow stars), showing it to be a good and principled choice. The sharp degradation in RMSE only occurs at very high FNR values (approaching 1.0), where the LOSS threshold becomes too relaxed to provide a meaningful signal.

\section{Varying the number of reference models}
\label{app:vary_k}

\begin{figure}[h]
    \centering
    \includegraphics[width=0.8\textwidth]{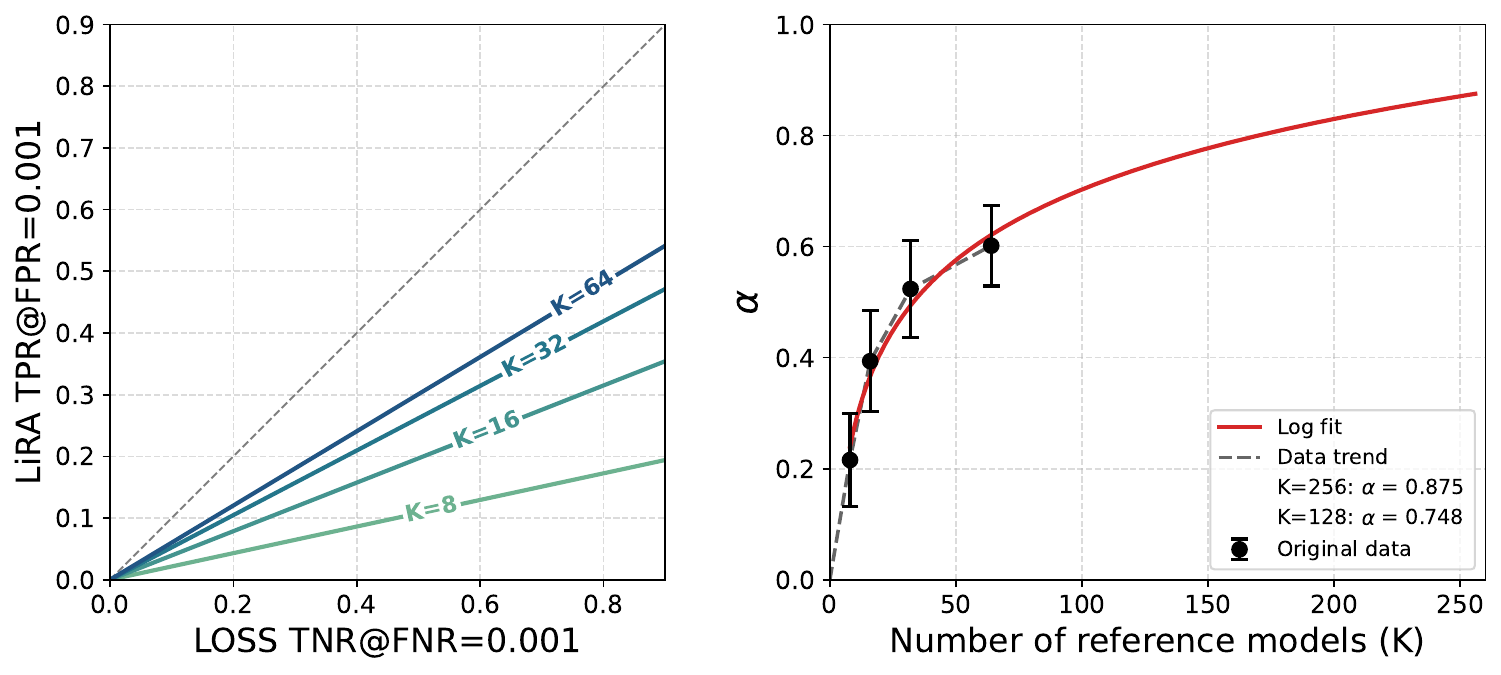}
    \caption{Performance of linear models for predicting LiRA with varying numbers of reference models (K). Left panel shows the correlation between LOSS TNR (x-axis) and LiRA TPR@$0.001$ (y-axis) across different K values, with linear regression fits shown for each condition. Right panel shows slope parameter ($\alpha$) for different K. The error bars show the 97.5\% confidence intervals obtained with bootstrapping.}
    \label{fig:different_shadows}
\end{figure}

The slope of the linear map from LOSS TNR to LiRA TPR is dependent on the strength of the attack being estimated. We instantiate LiRA with 8, 16, 32 and 64 reference models (half IN, half OUT) on each of our image setups, and fit a line $\mathrm{TPR}_{\mathrm{LiRA}} = \alpha_K \cdot \mathrm{TNR}_{\mathrm{LOSS}}$ for each $K$. Figure~\ref{fig:different_shadows} shows the fitted lines (left) and the slopes $\alpha_K$ with 97.5\% bootstrap confidence intervals (right).

The risk steadily increases with $K$ as the attack becomes more confident at identifying members. As noticed in previous work~\citep{carlini2022membership, hayes2025strongmembershipinferenceattacks}, TPR@low FPR experiences decreasing returns as the number of reference models increase. We can now model this with a logarithmic fit $\alpha_K = a \log K + b$ with $\alpha_{128} = 0.748$ and
$\alpha_{256} = 0.875$, giving practitioners a way to estimate risk against much stronger adversaries than they could afford to simulate. 

\section{Held-out evaluation}
\label{app:heldout}

In the main body, we report in-sample fit with bootstrap confidence intervals to quantify the uncertainty of that fit. To demonstrate our method also transfers to unseen architectures and datasets, we also report the cross-validated metrics here. We define two groupings: leave-one-architecture-out where we hold out each architecture in turn (10 folds), and leave-one-dataset-out where we do the same for each dataset (4 folds). In each fold, we refit the line on the retained configurations and predict the TPR on the held-out ones. We pool the RMSE over all held-out predictions. 

Table~\ref{tab:cv} reports the results. Under leave-one-architecture-out, the held-out RMSE is within $0.004$ of the in-sample fit at every FPR. We note that the RMSE is higher under leave-one-dataset-out. As CIFAR-100 defines the high end of the range, leaving it out forces the fit to extrapolate beyond the values it was trained on which is the expected cost of extrapolation. As we
show below, the ordering of models within each dataset is preserved.

\begin{table}[h]
\centering
\caption{Held-out RMSE under leave-one-group-out cross-validation, pooled over all
37 image configurations, compared with the in-sample fit.}
\label{tab:cv}
\begin{tabular}{lccc}
\toprule
 & FPR $=10^{-2}$ & FPR $=10^{-3}$ & FPR $=10^{-4}$ \\
\midrule
In-sample                   & 0.037 & 0.036 & 0.043 \\
Leave-one-architecture-out  & 0.038 & 0.038 & 0.047 \\
Leave-one-dataset-out       & 0.062 & 0.064 & 0.076 \\
\bottomrule
\end{tabular}
\end{table}

\paragraph{Within-dataset ranking.} Many of the workflows applicable to this work such as hyperparameter tuning, monitoring of trained models and architecture selection only require comparing a set of candidate models rather than estimating the absolute risk. This ranking does not depend on calibrating the slope. Table~\ref{tab:spearman} reports the rank correlation between LOSS TNR and LiRA TPR across architectures within each dataset. The correlation is above $0.91$ on every dataset, demonstrating that the LOSS TNR reliably ranks models by vulnerability, allowing practitioners to use it to compare models without any LiRA calibration.

\begin{table}[h]
\centering
\caption{Spearman rank correlation between LOSS TNR@FNR$=10^{-3}$ and LiRA TPR@FPR$=10^{-3}$
across architectures, computed within each dataset.}
\label{tab:spearman}
\begin{tabular}{lccc}
\toprule
Dataset & $n$ & $\rho$ & $p$ \\
\midrule
MNIST     & 9  & 0.950 & $8.8\times10^{-5}$ \\
CIFAR-10  & 10 & 0.915 & $2.0\times10^{-4}$ \\
CINIC-10  & 9  & 0.933 & $2.4\times10^{-4}$ \\
CIFAR-100 & 9  & 0.917 & $5.1\times10^{-4}$ \\
\bottomrule
\end{tabular}
\end{table}


\section{Fitting different functions}

So far we have estimated LiRA TPR with a simple linear model. We now study whether non-linear models lead to better risk estimates. Identifying non-members is indeed a matter of finding a good threshold to separate the tail of the loss distribution from the rest, while identifying members is dependent on reference models and is likely to be a more difficult task. We indeed find empirically that the LOSS TNR is typically higher than LiRA TPR at the same FPR. 

We thus compute goodness-of-fit metrics for convex functions such as a two-parameter polynomial, power-law, and exponential functions.

\begin{figure}[H]
    \centering
    \includegraphics[width=\linewidth]{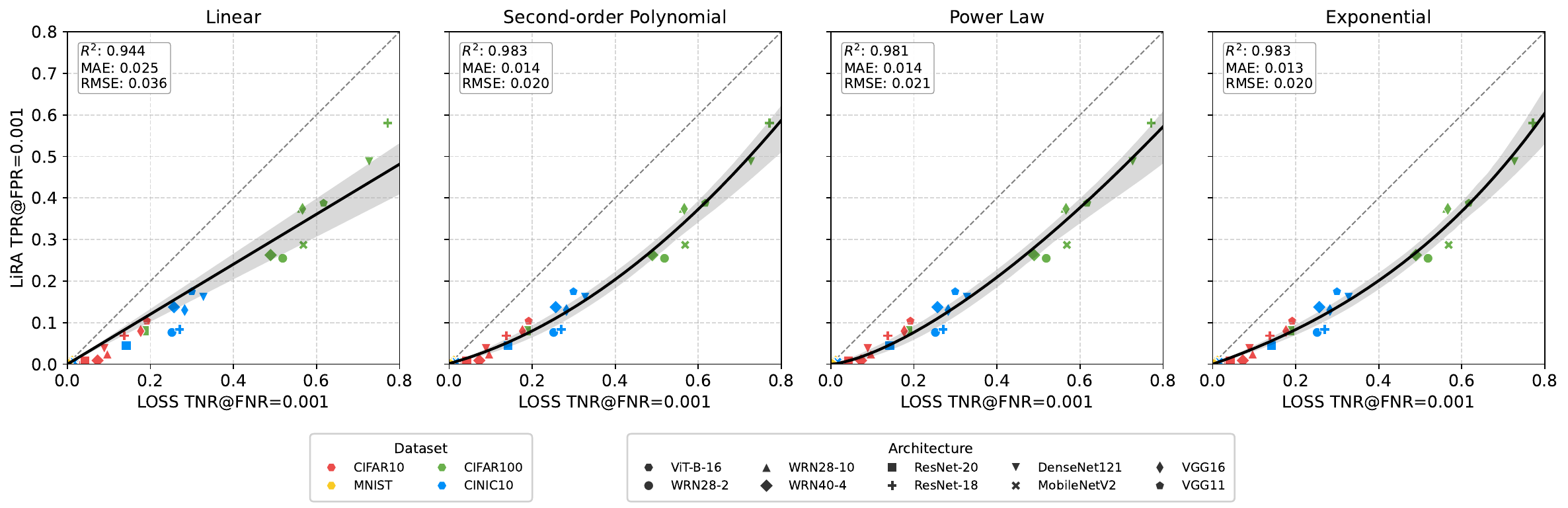}
    \caption{Comparison of regression models for predicting TPR@0.001 with LOSS TNR@0.001 evaluated on image recognition setups. The exponential function achieves the lowest RMSE of 0.02, compared to 0.036 achieved by the linear fit.}
    \label{fig:different_fits}
\end{figure}

Figure~\ref{fig:different_fits} shows that the exponential function $a(e^{bx}-1)$ achieves the best fit, outperforming the linear model. An exponential fit would imply that as LOSS TNR increases, member identification becomes easier as the model memorizes more difficult samples. Initial gains in LOSS TNR produce modest improvements in LiRA TPR, but these improvements accelerate as memorization increases. We nonetheless report the linear fit throughout the paper as it performs strongly and is more interpretable.

\section{Additional baselines}
\label{app:baselines}

Table~\ref{tab:baselines} reports additional reference-free statistics. Each statistic is mapped to LiRA TPR with a linear fit, with the origin fixed at the no-signal point (0.5 for AUC, the origin for tail statistics etc.).

\begin{table}[h]
\centering
\caption{The full set of reference-free
statistics evaluated, extending Table~\ref{tab:performance_metrics}. Fitting convention as in Table~\ref{tab:performance_metrics}.}
\label{tab:baselines}
\begin{tabular}{lccc}
\toprule
Metric & $R^2$ & RMSE & MAE \\
\midrule
RMIA (2 reference models) & 0.909 & 0.046 & 0.035 \\
LT-IQR AUC & 0.803 & 0.068 & 0.046 \\
LOSS AUC & 0.802 & 0.068 & 0.049 \\
KS distance & 0.787 & 0.070 & 0.052 \\
Train-test accuracy gap & 0.681 & 0.086 & 0.058 \\
Wasserstein-1 & 0.571 & 0.100 & 0.061 \\
Train-test loss gap & 0.562 & 0.101 & 0.061 \\
Quantile gap ($p_{50}$) & 0.503 & 0.108 & 0.072 \\
Quantile gap ($p_{90}$) & 0.456 & 0.113 & 0.068 \\
Quantile gap ($p_{99}$) & 0.204 & 0.136 & 0.086 \\
Partial AUC (FPR $\leq 10^{-3}$) & 0 & 0.153 & 0.120 \\
\midrule
\textbf{LOSS TNR} (ours) & \textbf{0.944} & \textbf{0.036} & \textbf{0.025} \\
\bottomrule
\end{tabular}
\end{table}

\section{Robustness across levels of generalization gaps}

Prior work has identified overfitting, typically measured by the train-test accuracy gap, as a key factor influencing membership inference vulnerability~\citep{yeom2018privacy, carlini2022membership}.

\begin{figure}[h]
  \begin{center}
    \includegraphics[width=0.45\textwidth]{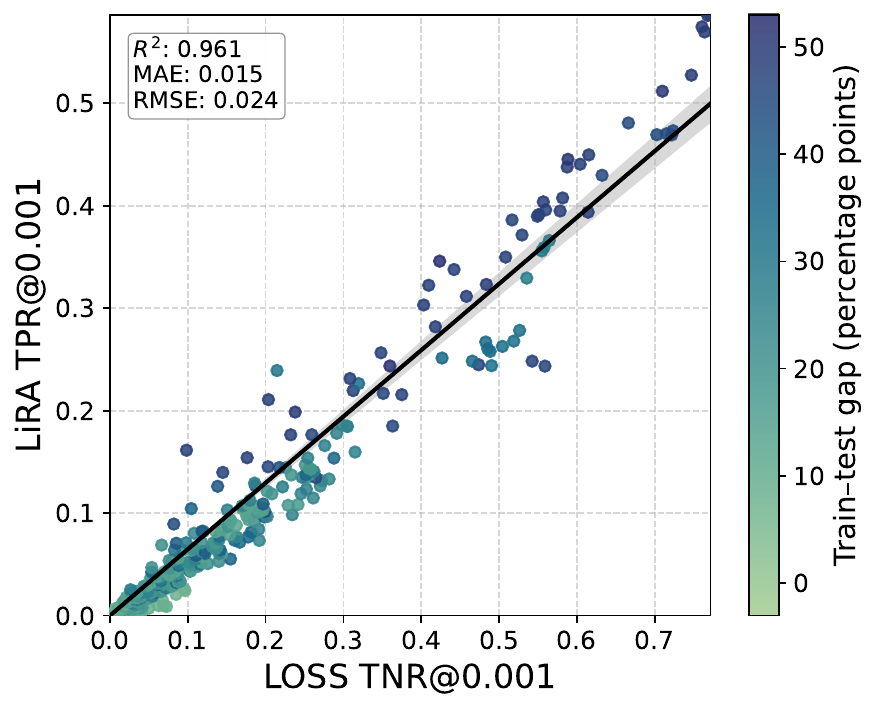}
  \end{center}
  \caption{LOSS TNR reliably predicts LiRA TPR across varying overfitting levels. }
  \label{fig:train_test_gap}
  \vspace{-10pt}
\end{figure}

We now study the effectiveness of our method for predicting the vulnerability of models with widely varying train-test gaps, ranging from near-zero (minimal overfitting) to over 50 percentage points (severe overfitting). Specifically, we treat intermediate training checkpoints from each setup in our main experiment as additional target models, instantiate LiRA against each target, and assess how well our method predicts its vulnerability. Figure~\ref{fig:train_test_gap} shows that LOSS TNR remains a reliable predictor of model vulnerability across this full range of generalization gaps.

This consistency demonstrates that LOSS TNR captures memorization patterns much richer than that of the overall generalization performance. While models with larger train-test gaps tend to have higher vulnerability (evidenced by the color gradient), TNR provides accurate risk estimates across a wide spectrum. 


\section{Predicting vulnerability to different attacks}
\label{app:attacks}
\begin{figure}[h]
    \centering
    \includegraphics[width=\linewidth]{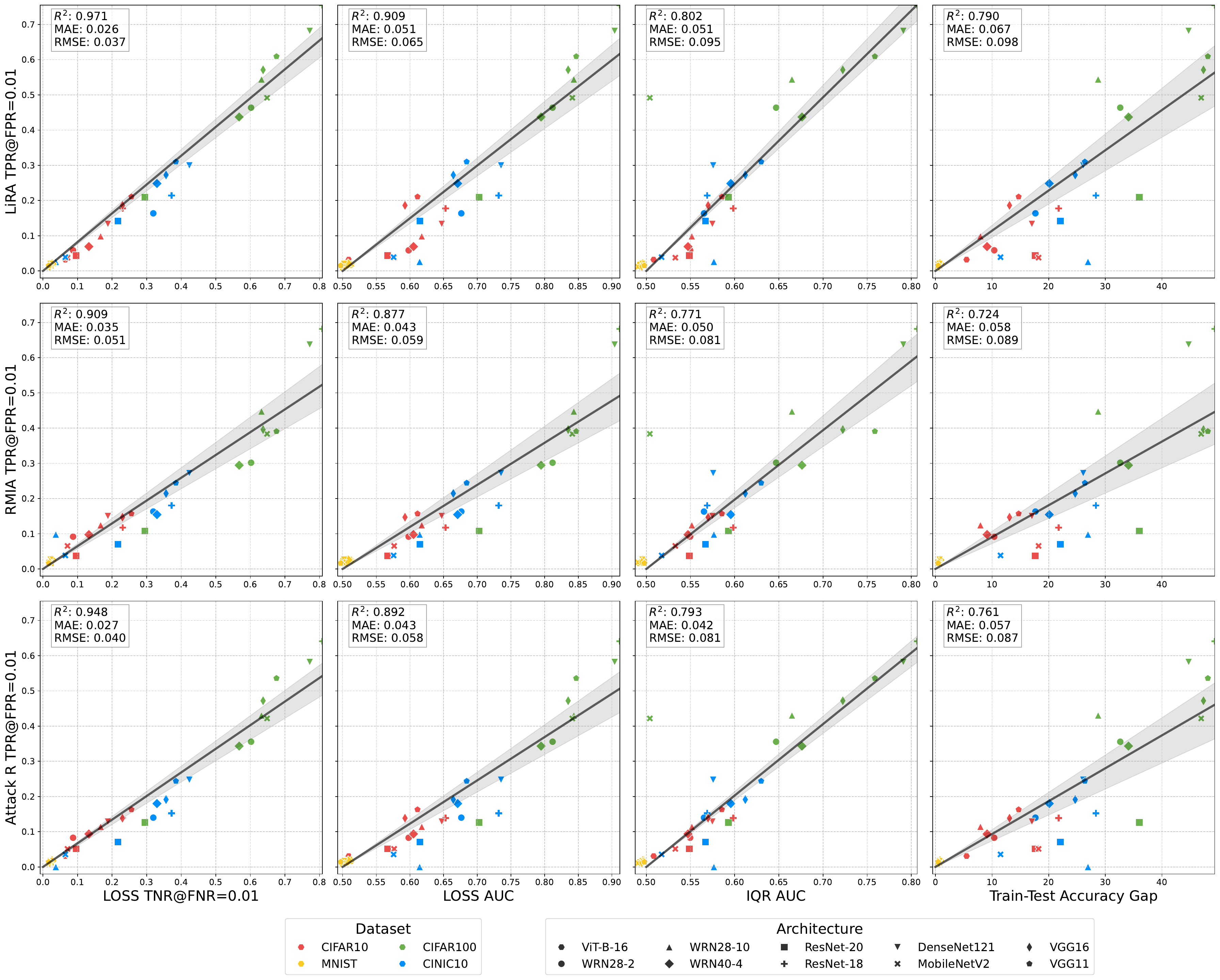}
    \caption{Comparison of different metrics for predicting the TPR at FPR=0.01 for LiRA, RMIA and Attack-R with 64 reference models. Each panel shows the relationship between a predictor metric (x-axis) and the attack TPR (y-axis) across all model-dataset combinations. Linear fits (solid lines) with 97.5\% confidence intervals (shaded regions) show goodness-of-fit.}
    \label{fig:attacks}
\end{figure}

To evaluate whether our method generalizes to other attacks within the same family, we estimate the TPR@0.01 for RMIA and Attack R using 64 reference models (for RMIA: 32 IN and 32 OUT; for Attack R: 64 OUT). Figure \ref{fig:attacks} demonstrates that TNR serves as an effective predictor for both RMIA and Attack R (two reference-model-based MIAs). Since these attacks share a common approach of using reference models to estimate counterfactuals for individual samples, it follows logically that TNR captures the same underlying signal across attacks in this family.

\section{Predicting vulnerability across different tasks}

\begin{figure}[h]
    \centering
    \includegraphics[width=0.4\linewidth]{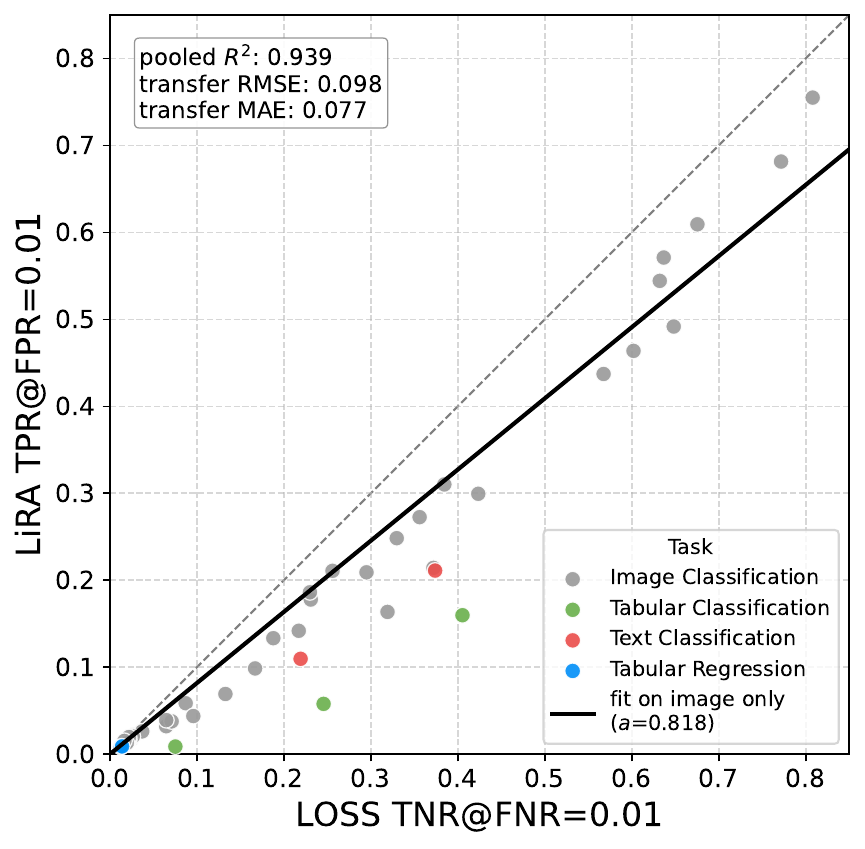}
    \caption{LOSS TNR successfully predicts model-level vulnerability to LiRA across tasks. Our metric achieves a pooled $R^2 = 0.939$. Fitting the prediction line on the image tasks alone and applying it to the non-image tasks gives $\mathrm{RMSE} = 0.098$, against $0.037$ within the image family.}
    \label{fig:new_tasks}
\end{figure}

To evaluate how our method performs for tasks beyond image classification, we instantiate it for 3 distinct tasks, specifically:
\begin{enumerate}
    \item \textbf{Text classification}
    \begin{enumerate}
        \item IMDb movie reviews dataset~\citep{maas2011learning}, TextCNN following~\citet{kim2014convolutional}. \\
        \item AGNews dataset~\citep{Zhang2015CharacterlevelCN}, TextCNN following~\citet{kim2014convolutional}. \\
    \end{enumerate}
    \item \textbf{Tabular classification}
    \begin{enumerate}
        \item US Census dataset~\citep{misc_adult_2}, 3-layer feedforward neural network with ReLU activation \\
        \item Texas hospital stays dataset~\footnote{https://www.dshs.texas.gov/THCIC/Hospitals/Download.shtm}, 4-layer feedforward neural network with Tanh activation \\
        \item Purchase dataset~\footnote{https://kaggle.com/c/acquire-valued-shoppers-challenge/data} as processed by~\citet{shokri2017membership}, 4-layer feedforward neural network with Tanh activation \\
    \end{enumerate}
    \item \textbf{Tabular regression}
    \begin{enumerate}
        \item California housing prices dataset, 3-layer feedforward neural network with ReLU activation~\citep{KELLEYPACE1997291} \\
        \item Bike sharing dataset~\citep{fanaee2014event}, 3-layer feedforward neural network with ReLU activation \\
    \end{enumerate}
\end{enumerate}

Figure~\ref{fig:new_tasks} shows that LOSS TNR remains a strong predictor of LiRA TPR, and, importantly, that the same fitted line is valid for these setups as for the image classification ones. However, the mapping moderately overestimates the vulnerability to these new task families which is the conservative approach for auditing. To achieve an even better fit, we recommend recalibrating the slope once per task family.

\section{Generalization to a second language-model family}
\label{app:pythia}

Section~\ref{sec:llms} evaluates five GPT-2 models. To test whether the weak-collapse behaviour and the LOSS AUC relationship are specific to that architecture and corpus, we also run experiments on Pythia models (8 models, 14M-2.8B parameters)~\citep{biderman2023pythiasuiteanalyzinglarge}, training on samples of the Pile~\citep{gao2020pile800gbdatasetdiverse}. 

\begin{figure}[h]
  \begin{center}
    \includegraphics[width=0.45\textwidth]{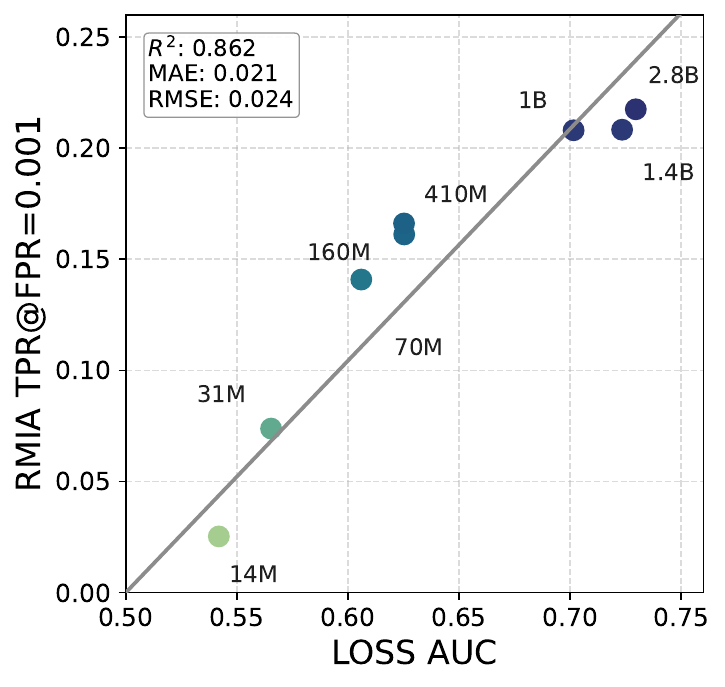}
  \end{center}
  \caption{LOSS AUC against RMIA TPR@FPR$=10^{-3}$ for eight Pythia models trained from scratch on the Pile.}
  \vspace{20pt}
  \label{fig:pythia_auc}
\end{figure}

Running LiRA on Pythia-scale models is prohibitively expensive, as it would require training many reference models. We thus use RMIA with $K=4$ as a lower-cost surrogate. This is motivated by our image model results, where RMIA, while less accurate than our loss-attack-based proxies, correlates well with LiRA as a model-level vulnerability estimator (Table~\ref{tab:performance_metrics}, $R^2 = 0.91$, with only two reference models). Using RMIA as a weak proxy for LiRA on Pythia, we observe the same monotonic relationship between LOSS AUC and attack performance that we see for the GPT-2 models: $R^2 = 0.862$, $\mathrm{RMSE} = 0.024$, Pearson $r = 0.941$
($p < 10^{-3}$) compared to $R^2 = 0.281$ and $\mathrm{RMSE} = 0.055$ when using LOSS TNR to estimate RMIA performance. This provides evidence that the weak-collapse behaviour is not specific to GPT-2, and that the LOSS AUC/LiRA relationship is likely to generalize to other model families.

\section{Computational cost}
\label{app:cost}

\begin{table}[h]
\centering
\caption{Cost of training the required number of reference models to run LiRA and RMIA for each image configuration, measured on a single NVIDIA A100 80GB. Each row reports the per-architecture cost for training on CIFAR-10 (25,000 members, batch size 256) at that architecture's training length. The last row spans all 10 architectures and 4 datasets from Section~\ref{sec:setup}. We also report the cost of our method, which only requires one
forward pass over the member and non-member sets of a model.}
\label{tab:cost}
\begin{tabular}{lrrrrr}
\toprule
Architecture & Params (M) & Epochs & LiRA ($K{=}64$) & RMIA ($K{=}2$) & Ours \\
 &  &  & (GPU-h) & (GPU-h) & (s) \\
\midrule
ResNet-18    & 11.2 & 200 &   5.8 & 0.18 &  0.5 \\
MobileNetV2  &  2.24 & 200 &  11.1 & 0.35 &  1.2 \\
ResNet-20    &  0.27 & 200 &  11.7 & 0.36 &  0.6 \\
VGG11        & 128.8 & 200 &  18.0 & 0.56 &  3.0 \\
WRN28-2      &  1.47 & 200 &  22.2 & 0.69 &  2.2 \\
VGG16        & 134.3 & 200 &  23.7 & 0.74 &  2.6 \\
WRN40-4      &  8.95 & 100 &  23.8 & 0.74 &  5.0 \\
DenseNet121  &  6.96 & 200 &  27.8 & 0.87 &  4.7 \\
WRN28-10     & 36.5 & 100 &  53.2 & 1.66 & 16.7 \\
\midrule
Range (37 configs) &  &  & 1.8--95.8 & 0.06--3.00 & 0.5--28.8 \\
\quad median       &  &  & 20.6 & 0.64 & 3.0 \\
\bottomrule
\end{tabular}
\end{table}

\begin{table}[h]
\centering
\caption{Estimated language-model cost. No training-time logs exist; figures are
$6ND$ for one training epoch and $2ND$ for the eval pass, at $40\%$ model-flop
utilisation on an A100 80GB (312~TFLOP/s, bfloat16). GPT-2 is trained on
$2^{19}$ sequences of 1024 tokens and attacked with LiRA at $K=127$. Pythia is
trained on a $50\%$ split of $2^{18}$ packed 256-token chunks and attacked with
RMIA at $K=4$ (2 IN, 2 OUT): four reference models, each on a random half of the
pool. Attack cost is the training of those references; ours is one forward pass
over members and non-members of a model that is already trained.}
\label{tab:cost-llm}
\begin{tabular}{llrr}
\toprule
Model & Attack & Attack cost & Ours (eval) \\
\midrule
GPT-2 10M    & LiRA ($K{=}127$) & 9 GPU-h & 3 min \\
GPT-2 140M   & LiRA ($K{=}127$) & 126 GPU-h & 40 min \\
GPT-2 302M   & LiRA ($K{=}127$) & 271 GPU-h & 1.4 GPU-h \\
GPT-2 604M   & LiRA ($K{=}127$) & 551 GPU-h & 2.9 GPU-h \\
GPT-2 1018M  & LiRA ($K{=}127$) & 932 GPU-h & 4.9 GPU-h \\
\midrule
Pythia 14M   & RMIA ($K{=}4$) & 0.03 GPU-h & 2 s \\
Pythia 31M   & RMIA ($K{=}4$) & 0.06 GPU-h & 5 s \\
Pythia 70M   & RMIA ($K{=}4$) & 0.13 GPU-h & 12 s \\
Pythia 160M  & RMIA ($K{=}4$) & 0.29 GPU-h & 27 s \\
Pythia 410M  & RMIA ($K{=}4$) & 0.73 GPU-h & 1.1 min \\
Pythia 1B    & RMIA ($K{=}4$) & 1.8 GPU-h & 2.8 min \\
Pythia 1.4B  & RMIA ($K{=}4$) & 2.5 GPU-h & 3.9 min \\
Pythia 2.8B  & RMIA ($K{=}4$) & 5.0 GPU-h & 7.6 min \\
\bottomrule
\end{tabular}
\end{table}

\FloatBarrier

\end{document}